\documentclass{article} % For LaTeX2e
\usepackage{iclr2023_conference,times}

\usepackage{amsmath,amsfonts,bm}

\def\eqref#1{equation~\ref{#1}}
\def\1{\bm{1}}

\DeclareMathAlphabet{\mathsfit}{\encodingdefault}{\sfdefault}{m}{sl}
\SetMathAlphabet{\mathsfit}{bold}{\encodingdefault}{\sfdefault}{bx}{n}

\usepackage{hyperref}
\usepackage{url}

\usepackage[utf8]{inputenc} % allow utf-8 input
\usepackage[T1]{fontenc}    % use 8-bit T1 fonts
\usepackage{hyperref}       % hyperlinks
\usepackage{url}            % simple URL typesetting
\usepackage{booktabs}       % professional-quality tables
\usepackage{amsfonts}       % blackboard math symbols
\usepackage{nicefrac}       % compact symbols for 1/2, etc.
\usepackage{microtype}      % microtypography
\usepackage{xcolor}         % colors
\usepackage{lipsum}
\usepackage{graphicx}
\usepackage{subfig}
\usepackage{amsthm}
\usepackage{algorithm}
\usepackage{algpseudocode}
\usepackage{amsmath}
\usepackage{pgfplots}
\usepackage{tikzscale}
\usepackage{wrapfig}

\pgfplotsset{compat=1.6}

\newcommand{\bo}[1]{\textbf{#1}}
\newcommand{\bw}[1]{\underline{#1}}

\def \betaTCVAE {$\beta$-TCVAE}
\def \factorVAE {FactorVAE}

\def \annealedVAE {AnnealedVAE}
\def \CVAE{ControlVAE}
\def \aaae {DAVA}

\def \dislib {\texttt{disentanglement-lib}}
\def \disc {PIPE}

\def \dsprites {\textit{DSprites}}
\def \abdsprites {\textit{AbstractDSprites}}
\def \shapes {\textit{Shapes3D}}
\def \mpitoy {\textit{Mpi3d Toy}}
\def \mpireal {\textit{Mpi3d Real}}
\def \celeba {\textit{CelebA}}
\def \compcars {\textit{CompCars}}

\theoremstyle{definition}
\newtheorem{definition}{Definition}[section]

\title{DAVA: Disentangling Adversarial Variational Autoencoder}

\author{%
 {Benjamin Estermann} \\
 ETH Zürich\\
 Switzerland \\
 \texttt{estermann@ethz.ch} \\
 \And
 {Roger Wattenhofer} \\
 ETH Zürich\\
 Switzerland \\
 \texttt{wattenhofer@ethz.ch} \\
}
\iclrfinalcopy % Uncomment for camera-ready version, but NOT for submission.
\begin{document}

\maketitle

\begin{abstract}
The use of well-disentangled representations offers many advantages for downstream tasks, e.g. an increased sample efficiency, or better interpretability.
However, the quality of disentangled interpretations is often highly dependent on the choice of dataset-specific hyperparameters, in particular the regularization strength.
To address this issue, we introduce \aaae{}, a novel training procedure for variational auto-encoders. \aaae{} completely alleviates the problem of hyperparameter selection.
We compare \aaae{} to models with optimal hyperparameters.
Without any hyperparameter tuning, \aaae{} is competitive on a diverse range of commonly used datasets.
Underlying \aaae{}, we discover a necessary condition for unsupervised disentanglement, which we call \disc{}.
We demonstrate the ability of \disc{} to positively predict the performance of downstream models in abstract reasoning.
We also thoroughly investigate correlations with existing supervised and unsupervised metrics. The code is available at \hyperlink{https://github.com/besterma/dava}{github.com/besterma/dava}.
\end{abstract}

\section{Introduction}
\begin{figure}[h!]
        \centering
        \includegraphics[width=\textwidth]{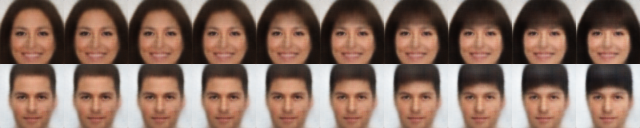}
        \caption{Latent traversals of a single latent dimension (hair fringes) of \aaae{} trained on \celeba{}. \aaae{} visibly disentangles the fringes from all other facial properties.} %like smile, hair length or gender
        
        \label{fig:celeba_fringes_smile}
    \end{figure}

    Real-world data tends to be highly structured, full of symmetries and patterns.
    This implies that there exists a lower-dimensional set of ground truth factors that is able to explain a significant portion of the variation present in real-world data.
    The goal of disentanglement learning is to recover these factors, so that changes in a single ground truth factor are reflected only in a single latent dimension of a model (see Figure \ref{fig:celeba_fringes_smile} for an example).
    Such an abstraction allows for more efficient reasoning \citep{van2019disentangled} and improved interpretability \citep{higgins2016beta}.
    It further shows positive effects on zero-shot domain adaption \citep{higgins2017darla} and data efficiency \citep{duan2019unsupervised, schott2022visual}.
    
    If the generative ground-truth factors are known and labeled data is available, one can train a model in a supervised manner to extract the ground-truth factors.
    What if the generative factors are unknown, but one still wants to profit from the aforementioned benefits for a downstream task?
    This may be necessary when the amount of labeled data for the downstream task is limited or training is computationally expensive.
    Learning disentangled representations in an unsupervised fashion is generally impossible without the use of some priors \citep{locatello2019challenging}.
    These priors can be present both implicitly in the model architecture and explicitly in the loss function \citep{tschannen2018recent}.
    An example of such a prior present in the loss function is a low total correlation between latent variables of a model \citep{chen2018isolating, kim2018disentangling}. 
    Reducing the total correlation has been shown to have a positive effect on disentanglement \citep{locatello2019challenging}.
    Unfortunately, as we show in more detail in this work, how much the total correlation should be reduced to achieve good disentanglement is highly dataset-specific.
    The optimal hyperparameter setting for one dataset may yield poor results on another dataset. 
    To optimize regularization strength, we need a way to evaluate disentanglement quality.

    So how can we identify well disentangled representations?
    Evaluating representation quality, even given labeled data, is no easy task.
    Perhaps as an example of unfortunate nomenclature, the often-used term ``ground-truth factor'' implies the existence of a canonical set of orthogonal factors.
    However, there are often multiple equally valid sets of ground truth factors, such as affine transformations of coordinate axes spanning a space, different color representations, or various levels of abstraction for group properties.
    This poses a problem for supervised disentanglement metrics, since they fix the ground truth factors for evaluating a representation and judge the models too harshly if they have learned another equally valid representation.
    Furthermore, acquiring labeled data in a practical setting is usually a costly endeavor. 
    The above reasons hinder the usability of supervised metrics for model selection.
    
    In this work, we overcome these limitations for both learning and evaluating disentangled representations.
    Our improvements are based on the following idea: 
    We define two distributions that can be generated by a VAE.
    Quantifying the distance between these two distributions yields a disentanglement metric that is independent of the specific choice of ground truth factors and reconstruction quality.
    The further away these two distributions are, the less disentangled the VAEs latent space is.
    We show that the similarity of the two distributions is a necessary condition for disentanglement.
    Furthermore, we can exploit this property at training time by introducing an adversarial loss into classical training of VAEs.
    To do this, we introduce a discriminator network into training and use the VAEs decoder as generator. 
    During training, we control the weight of the adversarial loss. 
    We adjust the capacity of the latent space information bottleneck accordingly, inspired by \citep{burgess2018understanding}.
    In this way, we allow the model to increase the complexity of its representation as long as it is able to disentangle.
    
    This dynamic training procedure solves the problem of dataset-specific hyperparameters and allows our approach to reach competitive disentanglement on a variety of commonly used datasets without hyperparameter tuning.

    Our contributions are as follows:

    \begin{itemize}
        \item We identify a novel unsupervised aspect of disentanglement called \disc{} and demonstrate its usefulness in a metric with correlation to supervised disentanglement metrics as well as a downstream task.
        \item We propose an adaptive adversarial training procedure (\aaae{}) for variational auto-encoders, which solves the common problem that disentanglement performance is highly dependent on dataset-specific regularization strength.
        \item We provide extensive evaluations on several commonly used disentanglement datasets to support our claims.
    \end{itemize}

\section{Related Work}
    \subsection{Model Architectures}
        The $\beta$-VAE by \citet{higgins2016beta} is a cornerstone model architecture for disentanglement learning. The loss function of the $\beta$-VAE, the evidence lower bound (ELBO), consists of a reconstruction term and a KL-divergence term weighted by $\beta$, which forces the aggregated posterior latent distribution to closely match the prior distribution.
        The KL-divergence term seems to promote disentanglement as shown in \citep{rolinek2019variational}. The $\beta$-TCVAE architecture proposed by \citet{chen2018isolating} further decomposes the KL divergence term of the ELBO into an index-code mutual information, a total correlation and a dimension-wise KL term. They are able to show that it is indeed the total correlation that encourages disentanglement and propose a tractable but biased Monte Carlo estimate. Similarly, the FactorVAE architecture \citep{kim2018disentangling} uses the density ratio trick with an adversarial network to estimate total correlation. The AnnealedVAE architecture \citep{burgess2018understanding} as well as ControlVAE \citep{shao2020controlvae} build on a different premise, arguing that slowly increasing the information bottleneck capacity of the latent space leads to the model gradually learning new latent dimensions and thus disentangling them. We will use a similar but optimized approach for \aaae{}. More recent promising approaches are presented by \citet{wei2021orthogonal} using orthogonal Jacobian regularization and by \citet{chen2021recursive} applying regulatory inductive bias recursively over a compositional feature space.

    \subsection{Introduction of Adversarial Training to VAEs}
        When combining autoencoders (AEs) with an adversarial setting, one can connect the adversarial network either on the latent space or on the output of the decoder.
        \citet{makhzani2015adversarial} proposed an adversarial AE (AAAE) that uses an adversarial discriminator network on the latent space of an AE to match its aggregated posterior to an arbitrary prior.
        There, the encoder of the AE acts as the generator of the GAN and is the only part of the AE that gets updated with respect to the discriminator's loss.
        This kind of training is strongly connected to VAE training, with the adversarial loss taking on the role of KL divergence in the classical VAE training objective but without the constraint of a Gaussian prior.
        The previously mentioned FactorVAE \citep{kim2018disentangling} implements an adversarial loss on the latent space to reduce total correlation.
        \citet{larsen2016autoencoding} proposed using a discriminator network with the decoder of the VAE acting as generator to improve the visual fidelity of the generated images, but with no focus on disentanglement. The difference to \aaae{} is that they used the discriminator on the real observations, while we propose a discriminator that only sees observations generated by the decoder of the VAE.
        \citet{zhu2020learning} introduce an recognition network based loss on the decoder that encourages predictable latent traversals. In other words, given a pair of images where all but one latent dimensions are kept constant, the recognition network should be able to predict in which latent dimension the change occurred. Applied on top of a baseline VAE model, this loss slightly improves the disentanglement performance of the baseline VAE.
        In a semi-supervised setting, \citet{carbajal2021disentanglement} and \citet{han2020disentangled} propose adversarial losses on the latent space of VAEs to disentangle certain parts of the latent space from information present in labels. 
        Unfortunately, such an approach does not work in an unsupervised setting with the goal of disentanglement of the complete latent space.
        
    \subsection{Measuring Disentanglement}
        All supervised metrics have in common that they are dependent on a canonical factorization of ground truth factors.
        Given access to the generative process of a dataset, the FVAE metric \citep{kim2018disentangling} can be used to evaluate a model. It first creates a batch of data generated by keeping aforementioned ground truth factor fixed and randomly sampling from the other ground truth factors.
        It then uses a majority vote classifier to predict the index of a ground truth factor given the variance of each latent dimension computed over said batch.
        Without access to the generative process, but given a number of fully labeled samples, one can use metrics like DCI Disentanglement by \citet{eastwood2018framework} and MIG by \citet{chen2018isolating}.
        While DCI in essence assesses if a latent variable of a model captures only a single ground truth factor, MIG evaluates if a ground truth factor is captured only by a single variable. As a result, DCI compared to MIG does not punish multi-dimensional representations of a single ground truth factor, for example the RGB model of color or a sine/cosine representation of orientation. 
        
        There exists a small number of unsupervised disentanglement metrics.
        The unsupervised disentanglement ranking (UDR) by Duan et al. \citep{duan2019unsupervised} evaluates disentanglement based on the assumption that a representation should be disentangled if many models, differently initialized but trained with the same hyperparameters, learn the same representation.
        To achieve this, they compute pairwise similarity (up to permutation and sign inverse) of the representations of a group of models.
        The score of a single model is the average of its similarity to all the other models in the group. 
        ModelCentrality (MC) \citep{lin2020infogan} builds on top of UDR by improving the pairwise similarity evaluation.
        A drawback is the high computational effort, as to find the optimal hyperparameter setting, multiple models need to be trained for each setting.
        UDR and MC do not assume any fixed set of ground truth factors.
        Nevertheless, a weakness of these approaches is that they do not recognize similarity of a group of models that each learn a different bijective mapping of the ground truth factors.
        The latent variation predictability (VP) metric by \citet{zhu2020learning} is based on the assumption that if a representation is well disentangled, it should be easy for a recognition network to predict which variable was changed given two input images with a change in only one dimension of the representation.
        An advantage of the VP metric is its ability to evaluate GANs as it is not dependent on the model containing an encoder.
        In comparison to our proposed metric, VP, UDR and MC are dependent on the size of the latent space or need to define additional hyperparameters to recognize inactive dimensions.
        \citet{estermann2020robust} showed that the UDR has a strong positive bias for low-dimensional representations, as low-dimensional representations are more likely to be similar to each other.
        One could see the same issue arise for the VP metric, as accuracy for the recognition network will likely be higher when there is only a low number of change-inducing latent dimensions available to choose from.

    \subsection{Applications of Disentangled Representations}
        The effects of disentangled representations on downstream task performance are manifold. \citet{van2019disentangled} showed that for abstract reasoning tasks, disentangled representations enabled quicker learning. We show in Section \ref{subsec:downstream_performance} that the same holds for our proposed metric.
        In work by \citet{higgins2017darla}, disentangled representations provided improved zero-shot domain adaption for a multi-stage reinforcement-learning (RL) agent. The UDR metric \citep{duan2019unsupervised} correlated well with data efficiency of a model based RL agent introduced by Watters et al. \citep{watters2019cobra}.
        Disentangled representations further seem to be helpful in increasing the fairness of downstream prediction tasks \citep{locatello2019fairness}. Contrary to previous qualitative evidence \citep{eslami2018neural, higgins2018scan}, \citet{montero2020role} present quantitative findings indicating that disentanglement does not have an effect on out-of-distribution (OOD) generalization. Recent work by \citet{schott2022visual} supports the claims of \citet{montero2020role}, concluding that disentanglement shows improvement in downstream task performance but not so in OOD generalization.

\section{Posterior Indifference Projection Equivalence}
    We first introduce the notation of the VAE framework, closely following the notation used by \citet{kim2018disentangling}. We assume that observations $x \sim \tilde{D}$ are generated by an unknown process based on some independent ground-truth factors. The goal of the encoder is to represent $x$ in a latent vector $z \in \mathbb{R}^d$. We introduce a Gaussian prior $p(z) = \mathcal{N}(0,\mathit{I})$ with identity covariance matrix on the latent space. The variational posterior for a given observation $x$ is then $q_\theta(z|x) = \prod_{j=1}^d\mathcal{N}(z_j|\mu_j(x), \sigma_j^2(x))$, where the encoder with weights $\theta$ outputs mean and variance. The decoder with weights $\phi$ projects from the latent space $z$ back to observation space $p_\phi(x|z)$. We can now define the distribution $q(z)$.
    \begin{definition}[EP]
        The empirical posterior (EP) distribution $q(z)$ of a VAE is the multivariate distribution of the latent vectors $z$ over the data distribution $\tilde{D}$. More formally, $q(z) = \mathbb{E}_{x \sim \tilde{D}}[q_{\theta}(z|x)]$. We can reconstruct an observation $x \sim \tilde{D}$ the following way: We sample $z \sim q_\theta(z|x)$ and then get the reconstruction $\hat{x} \sim p_{\phi}(x|z)$. We informally call observations generated by this process reconstructed samples and denote them as $\hat{x}$.
    \end{definition}
    The decoder is not constrained to only project from $q(z)$ to observation space. We can sample observations from the decoder by using different distributions on $z$. We define a particularly useful distribution.
    \begin{definition}[FP]
        The factorial posterior (FP) distribution $\Bar{q}(z)$ of a VAE is a multivariate distribution with diagonal covariance matrix. We define it as the product of the marginal EP distribution: $\Bar{q}(z) = \prod_{j=1}^{d}q(z_j)$. We can use the decoder to project $z \sim \Bar{q}(z)$ to observation space $\tilde{x} \sim p_{\phi}(x|z)$. We informally call observations created by this process generated samples and denote them as $\tilde{x}$. 
    \end{definition}
    
    We can now define the data distributions that arise when using the decoder to project images from either the EP or the FP.
    \begin{definition}[$\tilde{D}_{\text{EP}}, \tilde{D}_{\text{FP}}$]
        $\tilde{D}_{\text{EP}}$ is generated by the decoder projecting observations from the EP latent distribution $q(z)$, i.e. reconstructed samples. $\tilde{D}_{\text{EP}} = \mathbb{E}_{z \sim q(z)}[p_{\phi}(x|z)]$. $\tilde{D}_{\text{FP}}$ is generated by the decoder projecting observations from the FP distribution $\Bar{q}(z)$, i.e. generated samples. $\tilde{D}_{\text{FP}} = \mathbb{E}_{z \sim \bar{q}(z)}[p_{\phi}(x|z)]$.
    \end{definition}
    
    We can now define the core concept of this paper. We look at the similarity of two data distributions generated by the decoder, $\tilde{D}_{\text{EP}}$ and $\tilde{D}_{\text{FP}}$.
    \newpage
    \begin{definition}[PIPE]
        The posterior indifference projection equivalence (\disc{}) of a VAE represents the similarity of the data distributions $\tilde{D}_{\text{EP}}$ and $\tilde{D}_{\text{FP}}$.  In other words, \disc{} is a measure of the decoder's indifference to the latent distribution it projects from.\\
        $\mathit{PIPE}(\theta,  \phi) = \omega(\mathbb{E}_{z \sim q(z)}[p_{\phi}(x|z)], \mathbb{E}_{z \sim \bar{q}(z)}[p_{\phi}(x|z)]$, where $\omega$ is a general similarity measure.
    \label{def:pipe}
    \end{definition}
    Unsupervised disentanglement learning without any inductive biases is impossible as has been proven by \citet{locatello2019challenging}.
    Given a disentangled representation, it is always possible to find a bijective mapping that leads to an entangled representation.
    Without knowledge of the ground truth factors, it is impossible to distinguish between the two representations.
    We argue that \disc{} is necessary for a disentangled model, even when it is not sufficient.
    Suppose a model has learned a disentangled representation, but $\tilde{D}_{\text{EP}}$ is not equivalent to $\tilde{D}_{\text{FP}}$.
    There are two cases where this could happen.
    The first possibility is that the latent dimensions of the model are not independent. 
    This violates the independence assumption of the ground truth factors, so the model cannot be disentangled.
    The second possibility is that the model has not learned a representation of the ground truth factors and is generating samples that are not represented by the ground truth factors.
    This model would not be disentangled either. 
    We conclude that \disc{} is a necessary condition for disentanglement.

        \begin{figure}
        \centering
        \subfloat[Reconstructed samples.]{\includegraphics[width=0.6\textwidth]{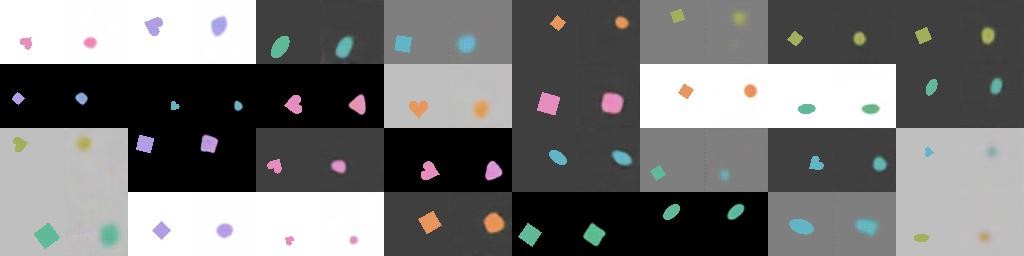}}
        \hfill
        \subfloat[Generated samples.]{\includegraphics[width=0.3\textwidth]{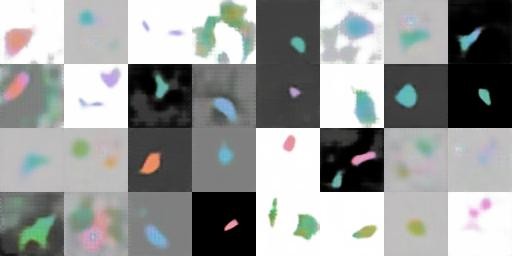}}
        \\
        \centering
        \subfloat[Reconstructed samples.]{\includegraphics[width=0.6\textwidth]{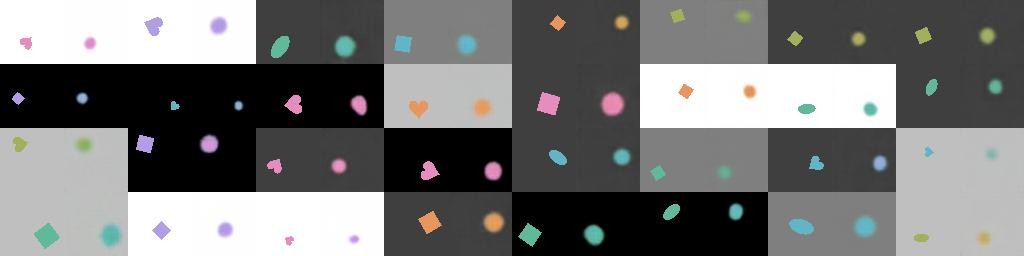}}
        \hfill
        \subfloat[Generated samples.]{\includegraphics[width=0.3\textwidth]{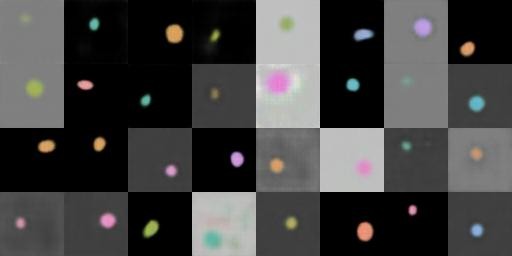}}
        \caption{Shown are two models of \dislib{} \citep{locatello2019challenging} with different disentanglement performance. The upper part shows a poorly disentangled model while the lower part shows a comparatively well disentangled model.
        In (a) and (c), odd columns show images sampled from the \abdsprites{} dataset, even columns show the corresponding model reconstruction. (b) and (d) show images generated by the decoder of each model given the factorial posterior distribution. While the upper model achieves better reconstructions (a), its generated samples (b) are visibly out of distribution. The bottom model ignores shape in its reconstructions (c), but its generated samples (d) capture the true data distribution more accurately.}
        \label{fig:metric/samples}
    \end{figure}
    
    In Figure \ref{fig:metric/samples} we show reconstructed and generated samples of two different models to support our chain of reasoning. 
    Generated samples of the entangled model (Figure \ref{fig:metric/samples} (b)) look visibly out of distribution compared to the reconstructed samples.
    Generated samples of the disentangled model (Figure \ref{fig:metric/samples} (d)) appear to be equivalent to the distribution of the reconstructed samples.

\section{PIPE Metric}
    We propose a way to quantify the similarity metric $\omega$ of \disc{} with a neural network and call this the \disc{} metric.
    Given a model $\mathcal{M}$ with encoder weights $\theta$, decoder weights $\phi$, and corresponding $\tilde{D}_{\text{EP}}, \tilde{D}_{\text{FP}}$ (see Appendix \ref{app:sampling_strategies} for details on how to sample from $\tilde{D}_{\text{EP}}$ and $tilde{D}_{FP}$).
    \begin{enumerate}
        \item Create a set of observations $S_{EP}$ by sampling from $\tilde{D}_{\text{EP}}$. Informally, these are the reconstructed samples.
        \item Create a set of observations $S_{FP}$ by sampling from $\tilde{D}_{\text{FP}}$. Informally, these are the generated samples.
        \item Randomly divide $S_{EP} \cup S_{FP}$ into a train and a test set.
        \item Train a discriminator network on the train set. %for $n_{train}$ steps. 
        \item Evaluate accuracy $acc$ of the discriminator on the test set.
        \item Since a random discriminator will guess half of the samples accurately, we report a score of $2\cdot(1 - acc)$, such that 1 is the best and 0 is the worst score.
    \end{enumerate}
    We train the discriminator network for $10,000$ steps to keep the distinction of $S_{EP}$ and $S_{FP}$ sufficiently difficult. We further use a uniform factorial distribution instead of FP for a slight improvement in performance. Details on the implementation can be found in Appendix \ref{app:metric_details}.
    
    \subsection{Relation to Existing Metrics}
\begin{figure}[h!]
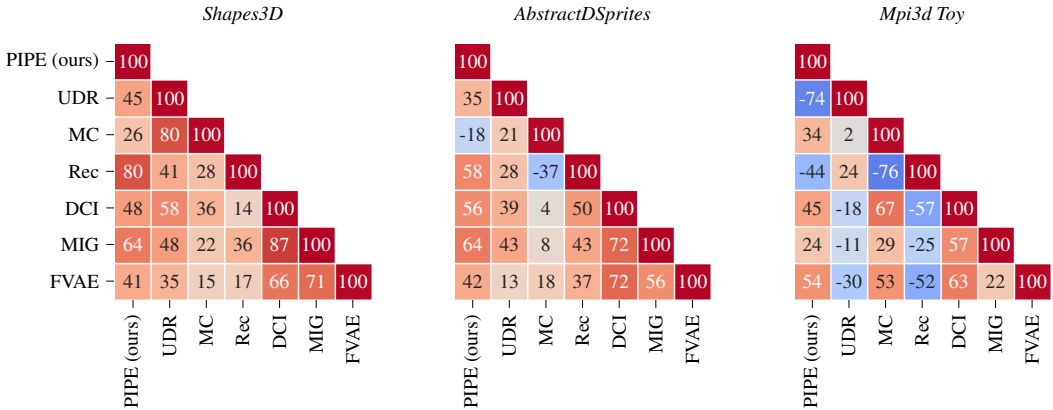

            \centering
            \centering
            \subfloat{\includegraphics[height=5.5cm]{figures/disc/disc_3DShapes_correlations.tikz}}
            \hfill
            \centering
            \subfloat{\includegraphics[height=5.5cm]{figures/disc/disc_AbstractDSprites_correlations.tikz}}
            \hfill
            \centering
            \subfloat{\includegraphics[height=5.5cm]{figures/disc/disc_Mpi3d_toy_correlations.tikz}}
            \caption{Spearman rank correlation between different metrics on three different datasets. Correlations take values in the range $[-1, 1]$, where 1 means perfect correlation, 0 means no correlation, and negative values mean anti-correlation. For better readability, we have multiplied all values by 100. \disc{}, UDR \citep{duan2019unsupervised} and MC \citep{lin2020infogan} are the corresponding unsupervised metrics. We include Reconstruction Loss Rec as a trivial unsupervised baseline. DCI, MIG and FVAE are the corresponding supervised metrics. Low or even negative correlations of our metric with UDR and MC show that our metric captures a different aspect of disentanglement.
            The correlation of our metric with supervised disentanglement metrics is mostly consistent across datasets. 
            We note that the correlation with supervised metrics on the \mpitoy{} dataset changes direction for both UDR and Rec, while MC has difficulties with \abdsprites{}.
            }
            \label{fig:metric/correlation_supervised}
        \end{figure}

        To classify the performance of the \disc{} metric, we evaluate correlations with supervised metrics on a diverse set of commonly used datasets. We namely consider \shapes{} \citep{3dshapes18}, \abdsprites{} \citep{van2019disentangled} and \mpitoy{} \citep{gondal2019transfer}.
        All datasets are part of \dislib{} \citep{locatello2019challenging}, which we used to train the models we evaluated our metric on. 
        More details on the implementation of the \disc{} metric and the evaluated models can be found in Appendix \ref{app:metric_details}.
        Results are displayed in Figure \ref{fig:metric/correlation_supervised}.
        They show that the \disc{} metric correlates positively with existing supervised metrics DCI, MIG and FVAE, surpassing the correlations of the unsupervised baselines.
        While the performance of UDR on \mpitoy{} and the performance of MC on \abdsprites{} is lacking, PIPE demonstrates consistent performance across all datasets. 
        Further, as opposed to UDR and MC, PIPE can be evaluated on a single model only, whereas UDR and MC are only able to evaluate sets of models.

    \subsection{Correlation with Downstream Task Performance}
    \label{subsec:downstream_performance}
        To further quantify the usefulness of the \disc{} metric, we analyze the predictive performance of the metric for an abstract reasoning downstream task. 
        For the downstream task, a VAE is trained to learn a disentangled representation of a dataset.
        The downstream model then only gets access to this representation then trying to solve the downstream task.
        \citet{van2019disentangled} evaluated different supervised disentanglement metrics in terms of predictive performance of accuracy in an abstract reasoning task on the datasets \shapes{} and \abdsprites{}. 
        They showed that good disentanglement is an indicator of better accuracy during the early stages on training. 
        We reproduce their experiment on a reduced scale by only considering up to 6,000 training steps.
        As can be seen in Figure \ref{fig:metric/downstream}, our metric is on par with supervised metrics and clearly outperforms the unsupervised baselines.
        More importantly this means that \disc{} is a positive predictor of accuracy of downstream models in a few-sample setting and is therefore a desirable property for unsupervised model selection.

    \begin{figure}[h!]
        \centering
        \includegraphics[height=4cm]{figures/downstream/Combined_correlations.tikz}
        \caption{Spearman rank correlation between different disentanglement metrics and downstream accuracy of the abstract visual reasoning task \citep{van2019disentangled} after 6,000 training steps for \factorVAE{} and \betaTCVAE{}. The same metrics as in Figure \ref{fig:metric/correlation_supervised} are evaluated. Correlation for unsupervised metrics are more sensitive to model architecture than for supervised metrics. \disc{} shows higher correlation with downstream performance than UDR and clearly outperforms the Rec baseline.}
        \label{fig:metric/downstream}
    \end{figure}

\section{Training Procedure for Variational Auto-encoders (\aaae{})}
    \begin{wrapfigure}{r}{0.5\textwidth}
        \centering
        \includegraphics[width=0.5\textwidth]{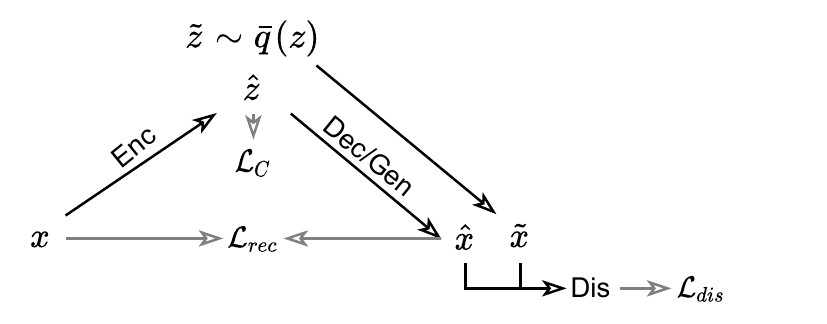}
        \caption{Overview of the composition of \aaae{}. A discriminator attached to the decoder of a VAE. The goal for the discriminator is to differentiate samples coming from the EP from those coming from the FP.}
        \label{fig:aaae_schematic}
    \end{wrapfigure}
    In the previous chapter we have established that PIPE is desirable for a model. 
    We now present how to encourage this property at training time by designing an adversarial training procedure.
    We train a discriminator and a VAE at the same time.
    The discriminator needs to differentiate $\tilde{D}_{\text{EP}}$ from $\tilde{D}_{\text{FP}}$, which is achieved with the following loss:
    \begin{equation}
        \nonumber 
        \begin{aligned}
                \mathcal{L}_{dis} = &\mathbb{E}_{\hat{x} \sim \tilde{D}_{\text{EP}}}[\mathrm{log}(Dis(\hat{x})] + \\
                                    &\mathbb{E}_{\tilde{x} \sim \tilde{D}_{\text{FP}}}[\mathrm{log}(1 - Dis(\tilde{x})) ].
        \end{aligned}
        \label{eq:loss_disc}
    \end{equation}
    The loss function for the VAE is compromised of a reconstruction term with the weighted negative objective of the discriminator:
    \begin{equation}
        \nonumber 
            \begin{aligned}
                &\mathcal{L}_{adv} = \mathcal{L}_{rec}  - \mu \mathcal{L}_{dis}.
            \end{aligned}
        \label{eq:loss_vae_disc}
    \end{equation}
    with reconstruction loss
    \begin{equation}
        \nonumber
        \mathcal{L}_{rec} = \mathbb{E}_{q_\theta(z|x)} [\mathrm{log} p_\phi(x | z)]
    \end{equation}

    While this works in practice, it faces the issue of the weight $\mu$ being very dataset-specific.
    It is also closely related to \factorVAE{}, with the difference that \factorVAE{} applies the adversarial loss to the latent space and to the encoder only.
    Let us consider an alternative approach. 
    Work by \citet{burgess2018understanding} looked at the KL-Divergence from an information bottleneck perspective.
    They proposed using a controlled latent capacity increase by introducing a loss term on the deviation of some goal capacity $C$ that increases during training:
    \begin{equation}
        \nonumber
        \mathcal{L}_{C} = |\mathrm{KL}(q(z|x)||p(z)) - C|
        \label{eq:loss_annealed}
    \end{equation}

    Such a loss provides a disentanglement prior complementary to the minimizing total correlation prior used in \betaTCVAE{} and \factorVAE{}.
    Unfortunately, it too depends on specific choice of hyperparameters, mainly the max capacity $C$ and the speed of capacity increase.
    We now demonstrate how to incorporate both the adversarial loss and the controlled capacity increase into a single loss function. We call this approach disentangling adversarial variational auto-encoder \aaae{} (also see Figure \ref{fig:aaae_schematic}):

    \begin{equation}
        \nonumber
        \mathcal{L}_{\mathit{DAVA}} = \mathcal{L}_{rec} - \gamma \mathcal{L}_{C} - \mu \mathcal{L}_{dis}
    \end{equation}
    
    The main strength of \aaae{} is its ability to dynamically tune $C$ and $\mu$ by using the accuracy of its discriminator. This yields a model that performs well on a diverse range of datasets.
    We now outline the motivation and the method for tuning $C$ and $\mu$ during training.
    
    \begin{wrapfigure}{R}{0.5\textwidth}
        \centering
        \input{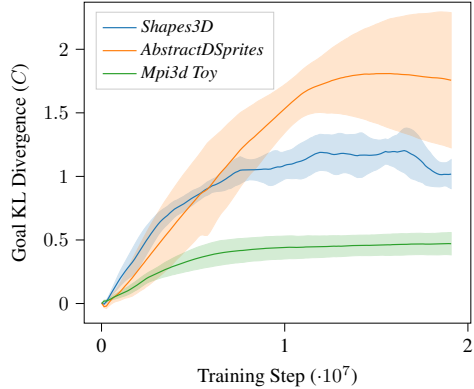}
        \caption{Information bottleneck capacity $C$ during training of \aaae{} for different datasets. Depending on the complexity and structure of the dataset as well as random initialization of the model, a different schedule of $C$ is necessary to achieve best performance. The shaded area denotes one standard deviation across 5 random seeds.}
        \label{fig:training/capacity}
        \vspace{-30pt}
    \end{wrapfigure}
    The goal of the controlled capacity increase in the \annealedVAE{} architecture is to allow the model to learn individual factors at a time into individual latent dimensions.
    The order in which they are learned corresponds to their respective contribution to the reconstruction loss.
    In \aaae{}, we want to encourage the model to \textit{learn new factors by increasing $C$ as long as it has learned a disentangled representation of the currently learned factors}.
    This is the case when the discriminator cannot distinguish $\tilde{D}_{\text{EP}}$ from $\tilde{D}_{\text{FP}}$.
    As soon as the accuracy of the discriminator increases, we want to stop increasing $C$ and increase the weight of the adversarial loss to ensure that the model does not learn any new factors of variation while helping it to disentangle the ones it has picked up into individual dimensions. 
    As can be seen in Figure \ref{fig:training/capacity}, this results in dataset-specific schedules of $C$ over the course of the training. 
    Algorithm \ref{algo:dava} in Appendix \ref{app:hyperparams} describes the training of \aaae{} in more detail.
    
    \subsection{Experiments}
    \label{subsec:dava_experiments}
    To quantify the performance of \aaae{}, we perform an extensive study including multiple model architectures and datasets.
    As a baseline, we include the best disentangling models according to \citep{locatello2019challenging}, namely the \betaTCVAE{} \citep{chen2018isolating} and \factorVAE{} \citep{kim2018disentangling}.
    We also include \annealedVAE{} \citep{burgess2018understanding} and \CVAE{} \citep{shao2020controlvae}.
    As discussed earlier, the performance of each of these approaches is highly dependent on the regularization strength.
    For this reason, we perform a grid-search over a range of values curated by roughly following the work of \citet{locatello2019challenging} and adjusting where necessary to ensure best possible performance.
    The exact hyperparameters considered can be found in appendix \ref{app:hyperparams}.
    The training of \aaae{} does not rely on such dataset-specific regularization, therefore \textbf{all hyperparameters} for the training of \aaae{} were kept \textbf{constant} across \textbf{all} considered datasets.
    All other hyperparameters closely follow \citep{locatello2019challenging} and were kept consistent across all architectures (baseline and \aaae{}) and are also reported in appendix \ref{app:hyperparams}.
    We run 5 different random seeds for each configuration to determine statistical significance.
    We provide summarized results for the most informative datasets in Tables \ref{tab:results_shapes}, \ref{tab:results_abdsprites} and \ref{tab:results_mpitoy}.
    For the baselines we report the performance of the mean hyperparameter setting according to the average score of all considered metrics.
    The reader is advised to consult Appendix \ref{app:complete_results} for complete results of all evaluated datasets.

        \begin{table}[h]
      \caption{Results \shapes{}}
      \label{tab:results_shapes}
        \centering
        \scriptsize
        \begin{tabular}{ l c c c c }
            \toprule
            Architecture &                      MIG &                   DCI &                   FVAE &              PIPE \\
            \midrule
            \betaTCVAE{} &                 0.26±0.06 &             0.50±0.03 &             0.76±0.03 &         0.11±0.03\\
            \factorVAE{} &                 0.20±0.10 &             0.41±0.07 &             0.78±0.04 &         0.22±0.04\\
            \annealedVAE{} &               0.52±0.02 &             0.70±0.02 &             \bo{0.92±0.01} &    0.26±0.03\\
            \CVAE{} &                      0.07±0.03 &             0.17±0.04 &             0.68±0.07 &         0.02±0.00 \\
            Ours &                         \bo{0.62±0.05} &        \textbf{0.78±0.03} &    0.82±0.03 &         \textbf{0.61±0.04} \\
            \bottomrule
        \end{tabular}
    \end{table}
    
    \begin{table}[h]
      \caption{Results \abdsprites{}}
      \label{tab:results_abdsprites}
        \centering
        \scriptsize
        \begin{tabular}{ l c c c c }
            \toprule
            Architecture &                      MIG &                   DCI &                   FVAE &                  PIPE \\
            \midrule
            \betaTCVAE{} &                 0.12±0.01 &             0.18±0.01&              0.45±0.02 &             0.19±0.01\\
            \factorVAE{} &                 0.12±0.02 &             0.19±0.02 &             0.55±0.03 &             0.20±0.03\\
            \annealedVAE{} &               0.15±0.01 &             0.21±0.01 &             0.54±0.02 &             0.18±0.01\\
            \CVAE{} &                      0.04±0.02 &             0.07±0.01 &             0.42±0.02 &             0.06±0.01 \\
            Ours &                         \textbf{0.23±0.04} &    \textbf{0.27±0.05} &    \textbf{0.67±0.05} &    \textbf{0.35±0.03} \\
            \bottomrule
        \end{tabular}
    \end{table}
    
    \begin{table}[h]
      \caption{Results  \mpitoy{}}
      \label{tab:results_mpitoy}
        \centering
        \scriptsize
        \begin{tabular}{ l c c c c }
            \toprule
             Architecture &                         MIG &                   DCI &                   FVAE &                  PIPE \\
            \midrule
            \betaTCVAE{} &                     0.11±0.02 &             0.23±0.01 &             0.39±0.02 &             0.09±0.01\\
            \factorVAE{} &                     0.02±0.01 &             0.13±0.00 &             0.38±0.02 &             (0.99±0.10)\\
            \annealedVAE{} &                   0.07±0.03 &             0.23±0.02 &             \textbf{0.50±0.02} &    0.02±0.01\\
            \CVAE{} &                          0.04±0.01 &             0.17±0.02 &             0.43±0.04 &             0.03±0.01\\
            Ours &                             \bo{0.12±0.09} &        \bo{0.30±0.03} &        0.41±0.04 &             \bo{0.21±0.03} \\
            \bottomrule
        \end{tabular}
    \end{table}

    Appendix \ref{app:complete_results} also contains the results of the best performing hyperparameter choice for each baseline model, demonstrating their variance across datasets.
    Note that the PIPE scores of \factorVAE{} in Table \ref{tab:results_mpitoy} are exploiting a design decision discussed in depth in Appendix \ref{subsec:app/reconstruction_loss}.
    The decoders of these models mapped all latent values to a single image, therefore we exclude their scores.
    The results showcase the strong consistent performance of \aaae{} across different datasets.
    Compared to the baselines, \aaae{} is able to deliver improvements of up to 50\% for the supervised metrics, and over 100\% for \disc{}.
    This allows \aaae{} to be used in practice as consistent performance can be expected.

    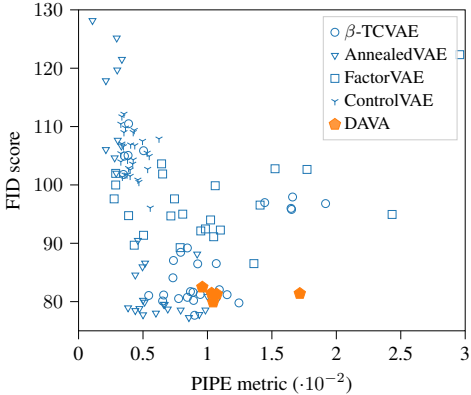
\begin{wrapfigure}{R}{0.5\textwidth}
            \centering
            % This file was created with tikzplotlib v0.10.1.
\begin{tikzpicture}[scale=0.75]

\definecolor{darkgray176}{RGB}{176,176,176}
\definecolor{darkorange25512714}{RGB}{255,127,14}
\definecolor{steelblue31119180}{RGB}{31,119,180}
\definecolor{lightgray204}{RGB}{204,204,204}

\begin{axis}[
legend style={
  fill opacity=0.8,
  draw opacity=1,
  text opacity=1,
  draw=lightgray204,
  font=\small,
},
legend cell align={left},
tick align=outside,
tick pos=left,
x grid style={darkgray176},
xlabel={\disc{} metric ($\cdot 10^{-2}$)},
xmin=0, xmax=0.03,
xtick style={color=black},
xticklabels={0, 0, 0.5, 1, 1.5, 2, 2.5, 3},
scaled ticks=false,
y grid style={darkgray176},
ylabel={FID score},
ymin=75, ymax=130,
ytick style={color=black}
]
\addplot [draw=steelblue31119180, fill=steelblue31119180, mark=o, only marks]
table{%
x  y
0.00504161331626096 105.864536131208
0.00384122919334162 105.057374797728
0.00348111395646611 101.7951475713
0.003881241997439 110.484608037535
0.00356113956466064 104.941754304474
0.0109234955185658 82.0426569951194
0.00652208706786173 80.1082679054034
0.0106434058898848 80.5747224935577
0.0054417413572343 81.0516206189849
0.00868277848911636 81.7221818692583
0.0124439820742637 79.7808843957021
0.00900288092189494 77.6652969603326
0.00660211267605626 81.1364898668757
0.00892285531370041 80.2088961952244
0.00844270166453276 80.7364336836594
0.00944302176696543 81.2258143957471
0.00732234314980773 84.0842103225162
0.0115236875800255 81.1927269876082
0.00888284250960303 81.6149718951355
0.00776248399487822 80.5254811726524
0.00924295774647876 86.4737327146006
0.00792253521126751 88.4558281955885
0.00736235595390511 87.0486541832381
0.0106834186939821 86.5225143015089
0.00844270166453276 89.1994925831022
0.0165252880921896 95.7965772460511
0.016485275288092 95.9846853888395
0.0166053137003839 97.9296881128553
0.0144446222791292 96.9694353962724
0.0191661331626118 96.8008801653797
};
\addlegendentry{\betaTCVAE{}}
\addplot [
  draw=steelblue31119180,
  fill=steelblue31119180,
  mark options={rotate=180},
  mark=triangle,
  only marks
]
table{%
x  y
0.00500160051216403 77.7411889921237
0.0060019206145967 78.0263111210801
0.00512163892445572 79.9004876908625
0.00440140845070425 78.4615095541371
0.00692221510883484 78.7265109584223
0.00660211267605626 79.5445389699268
0.00472151088348283 78.8124937629195
0.0100032010243276 80.8965685973258
0.00384122919334162 78.9358051779718
0.00500160051216403 80.2268517373538
0.00440140845070425 84.5538086663187
0.00496158770806665 85.9609734551629
0.00460147247119069 90.4902795110881
0.00920294494238161 88.1480922575823
0.0051616517285531 86.5987216730174
0.00292093469910371 102.010213515807
0.00332106274007682 106.660947270845
0.00212067861715748 106.053450205848
0.00304097311139562 107.651982989728
0.00280089628681157 104.633137417048
0.00296094750320086 125.168349001288
0.00108034571062743 128.163131605438
0.0033610755441742 121.510619458362
0.00300096030729824 119.689071879056
0.00212067861715748 117.852909153557
0.0093629961587709 77.7850932284467
0.00984314980793832 78.5576049354185
0.00856274007682445 77.2654261368384
0.00792253521126751 78.5475356017477
0.00668213828425102 79.5048985526376
};
\addlegendentry{\annealedVAE{}}
\addplot [draw=steelblue31119180, fill=steelblue31119180, mark=square, only marks]
table{%
x  y
0.00984314980793832 92.4551982338197
0.00788252240717036 89.252370325114
0.00432138284250949 89.6710050587176
0.0080825864276568 94.9920821615701
0.00504161331626096 91.3699904433328
0.0136043533930856 86.5162058359946
0.00948303457106281 92.114841593646
0.0104833546734955 91.1208238411752
0.00276088348271464 97.6129253445088
0.0243277848911652 94.9413156544616
0.00644206145966697 103.610758889181
0.00744238156209964 97.6126656879004
0.0110035211267605 92.2653394677357
0.003881241997439 94.736330819681
0.0102432778489117 93.9997570639363
0.00716229193341866 94.6854914982912
0.0106033930857874 99.8641363619077
0.00288092189500633 99.9874198019074
0.0140845070422535 96.5449768187954
0.00288092189500633 101.979361589078
0.0152448783610757 102.773436775424
0.0177256722151087 102.655036280264
0.0649007682458387 103.403270746293
0.00652208706786173 101.875914171535
0.0799455825864277 110.13313435993
0.166733354673495 133.350796075684
0.0645006402048656 114.143295305905
0.0469750320102431 110.178960913911
0.0540572983354672 308.859305795704
0.0295694622279128 122.32101416337
};
\addlegendentry{\factorVAE{}}
\addplot [
  draw=steelblue31119180,
  fill=steelblue31119180,
  mark=Mercedes star flipped,
  only marks
]
table{%
x  y
0.003881241997439 102.84967690705
0.00400128040973091 101.511775195906
0.00436139564660687 106.75640736009
0.00380121638924469 104.274582208531
0.00556177976952621 96.067254875107
0.00460147247119069 100.862671417382
0.00420134443021758 103.592271666208
0.00424135723431496 104.307144742858
0.00400128040973091 102.587136345439
0.00468149807938545 100.467655186473
0.00348111395646611 106.584217990898
0.00404129321382829 106.881520816848
0.00328104993597944 103.982201395277
0.00496158770806665 107.509392794869
0.00360115236875802 101.73532299616
0.00548175416133168 104.920817048116
0.00328104993597944 105.38523259288
0.00432138284250949 109.367265548236
0.0033610755441742 107.044892943858
0.00376120358514731 101.739721394491
0.00352112676056349 112.152469125145
0.00532170294494239 105.46895418403
0.00352112676056349 109.001934038745
0.0036411651728554 109.77936972481
0.00340108834827135 101.359312053009
0.00424135723431496 109.007848746396
0.0062419974391803 107.896959804323
0.0033610755441742 111.704758803519
0.00328104993597944 110.442960410088
0.00536171574903954 102.698966368019
};
\addlegendentry{\CVAE{}}
\addplot [draw=darkorange25512714, fill=darkorange25512714, mark=pentagon*,
mark size=3pt, only marks]
table{%
x  y
0.0171654929577465 81.3940907597253
0.0103233034571062 81.4416526267365
0.0104433418693981 79.9238072144411
0.0107234314980795 81.2707552986579
0.0096030729833545 82.481068116066
};
\addlegendentry{\aaae{}}
\end{axis}

\end{tikzpicture}
            \caption{The FID score \citep{heusel2017gans} measures the distance between two different distributions of images and is commonly used in GAN literature to evaluate the quality of the generator. A low FID indicates that a model generates visually convincing images. We evaluate the FID of image samples from $\tilde{D}_{\text{FP}}$ of the respective models against the original images. We find that \aaae{} achieves a competitive FID with comparatively good \disc{}.  We have removed a small number of poorly performing outliers from the baseline models to improve readability.}
            \vspace{-40pt}
            \label{fig:real_world/celeba_fid_vs_disc}
    \end{wrapfigure}
    
    We further validate \aaae{} on three real-world datasets. CelebA \citep{liu2015faceattributes} is dataset commonly used in the disentanglement literature. It compromises of over 200,000 images of faces of celebrities. As shown in Figure \ref{fig:real_world/celeba_fid_vs_disc}, \aaae{} achieves a competitive FID  while still having a comparatively high \disc{} metric score.
    This means that \aaae{} accomplishes a similar visual fidelity of its generated samples as the best baseline methods while better disentangling its latent space.
    Results for datasets \mpireal{} \citep{gondal2019transfer} and \compcars{} \citep{yang2015large} can be found in Appendix \ref{app:complete_results} and \ref{app:compcars_results}, respectively.

\section{Conclusions}
\label{sec:conclusions}
In this work, we presented a new way to understand and formalize disentangled representations from an unsupervised point of view.
The main idea is that the distribution of reconstructed samples of a disentangled VAE should be similar to the distribution of generated samples.
We quantify this notion in a new unsupervised disentanglement metric called \disc{} metric for evaluating disentangled representations.
Without access to ground truth, \disc{} provides results consistent with supervised metrics. Furthermore, we apply the same idea to \aaae{}, a novel VAE training procedure.
\aaae{} can self-tune its regularization strength and produces results competitive with existing models that require more dataset-specific hyperparameter tuning.
Robust unsupervised disentanglement performance is an important requirement for practical applicability.
With \aaae{}, we provide a method that can be used by practitioners with relatively low computational requirements.

\clearpage

\bibliography{iclr2023_conference}
\bibliographystyle{iclr2023_conference}

\clearpage
\appendix
\section{Datasets}
    We provide a short summary of each dataset, including samples and the respective data generating ground truth factors.
    \subsection{Shapes3D}
        The \shapes{} dataset \citep{3dshapes18} includes the following ground truth factors: floor hue, wall hue, object hue, scale, shape and orientation.
        \begin{figure}[h!]
            \centering
            \includegraphics[width=\textwidth]{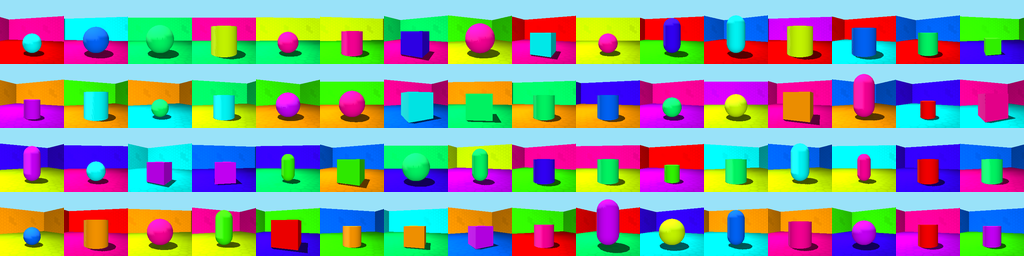}
            \caption{Samples from \shapes{}.}
        \end{figure}
    \subsection{DSprites}
        The \dsprites{} dataset \citep{dsprites17} includes the following ground truth factors: shape, scale, orientation, x-position and y-position.
        \begin{figure}[h!]
            \centering
            \includegraphics[width=\textwidth]{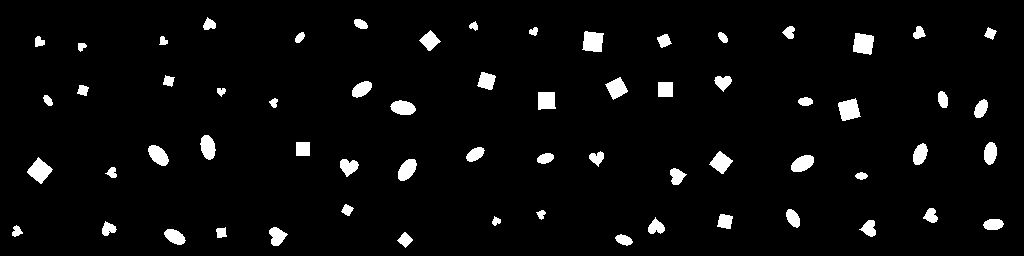}
            \caption{Samples from \dsprites{}.}
        \end{figure}
    \subsection{AbstractDSprites}
        The \abdsprites{} dataset \citep{van2019disentangled} includes the following ground truth factors: background color, object color, shape, scale, x-position and y-position.
        \begin{figure}[h!]
            \centering
            \includegraphics[width=\textwidth]{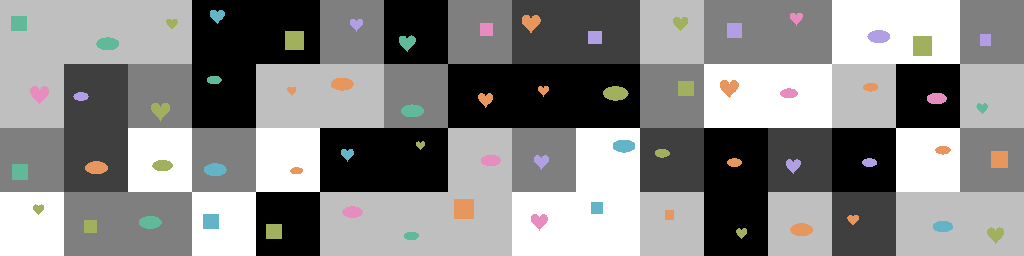}
            \caption{Samples from \abdsprites{}.}
        \end{figure}
    \subsection{Mpi3d Toy}
        The \mpitoy{} dataset \citep{gondal2019transfer} includes the following ground truth factors: object color, object shape, object size, camera height, background color, first degree of freedom, second degree of freedom. It consists of low-quality renders of the experimental setup.
        \begin{figure}[h!]
            \centering
            \includegraphics[width=\textwidth]{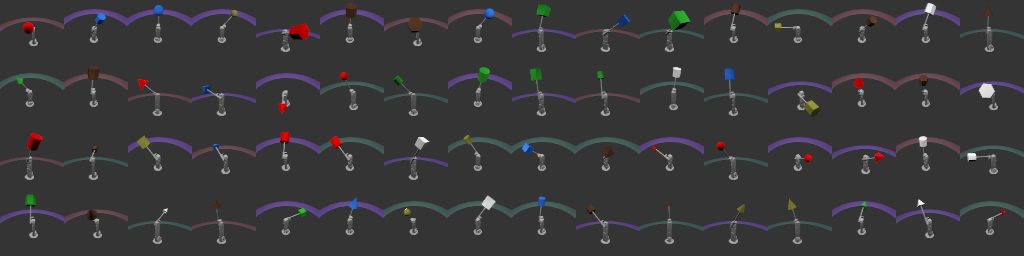}
            \caption{Samples from \mpitoy{}.}
        \end{figure}
    \subsection{Mpi3d Real}
        The \mpireal{} dataset \citep{gondal2019transfer} includes the same ground truth factors as \mpitoy{}. It consists of real pictures of an experimental setup.
        \begin{figure}[h!]
            \centering
            \includegraphics[width=\textwidth]{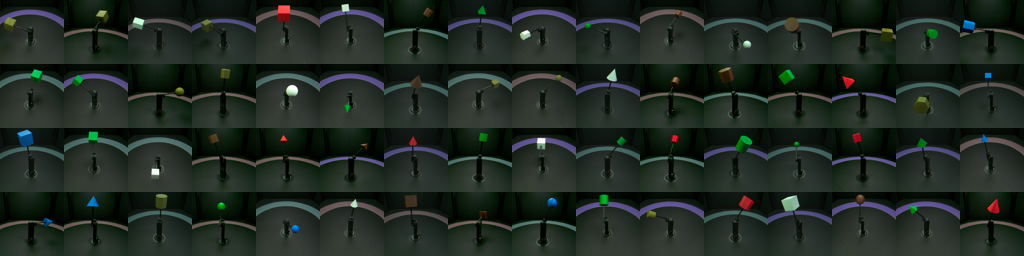}
            \caption{Samples from \mpireal{}}
        \end{figure}
    \subsection{Smallnorb}
        The \textit{Smallnorb} dataset \citep{lecun2004learning} includes the following ground truth factors: object type, elevation, azimuth, lighting condition.
        \begin{figure}[h!]
            \centering
            \includegraphics[width=\textwidth]{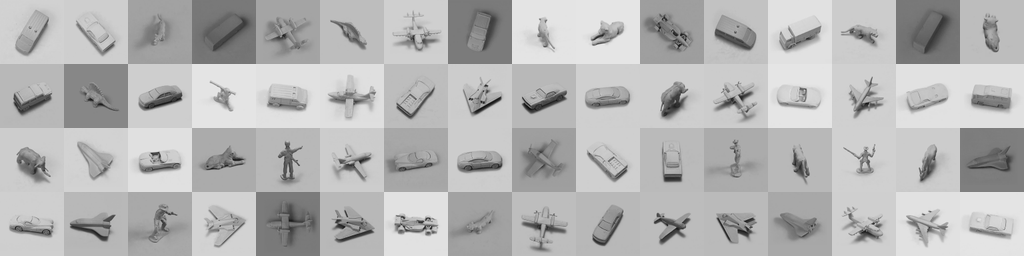}
            \caption{Samples from \textit{Smallnorb}.}
        \end{figure}
    \subsection{NoisyDSprites}
        The \textit{NoisyDSprites} dataset \citep{locatello2019challenging} includes the same ground truth factors as \dsprites{}. Instead of the background being black, it consists of random noise.
        \begin{figure}[h!]
            \centering
            \includegraphics[width=\textwidth]{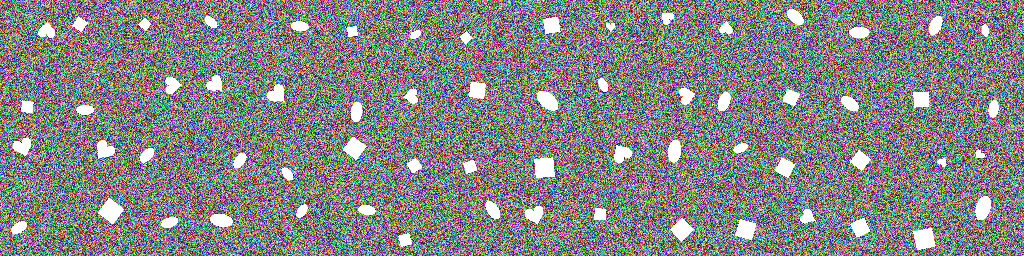}
            \caption{Samples from \textit{NoisyDSprites}}
        \end{figure}
    \subsection{Cars3D}
        The \textit{Cars3D} dataset \citep{reed2015deep} includes the following ground truth factors: elevation, azimuth, object type.
        \begin{figure}[h!]
            \centering
            \includegraphics[width=\textwidth]{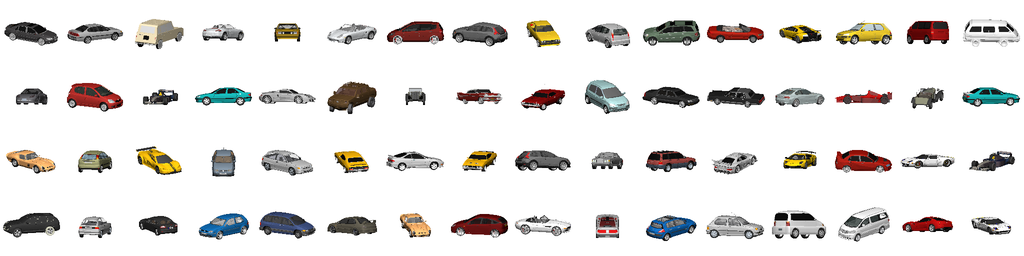}
            \caption{Samples from \textit{Cars3D}.}
        \end{figure}
    \subsection{CelebA}
        The \celeba{} dataset \citep{liu2015faceattributes} includes multiple images of over 10.000 identities with a rich set of facial properties, poses and background clutter.
        \begin{figure}[h!]
            \centering
            \includegraphics[width=\textwidth]{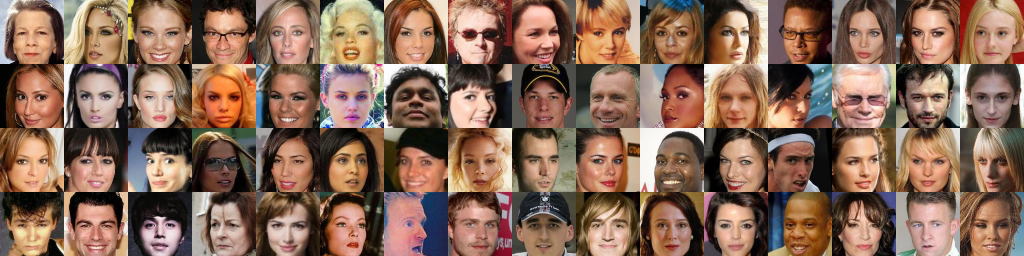}
            \caption{Samples from \celeba{}.}
        \end{figure}
    \newpage
    \subsection{CompCars}
        \label{app:compcars}
        The \compcars{} dataset \citep{yang2015large} contains images of 163 car makes with 1.716 car models, totalling to 136.726 images. The images scraped from the web and are composed of the car in different poses, eg. front and side view, with diverse backgrounds. It is uncleaned and might contain occlusions and highly correlated ground truth factors.
        \begin{figure}[h!]
            \centering
            \includegraphics[width=\textwidth]{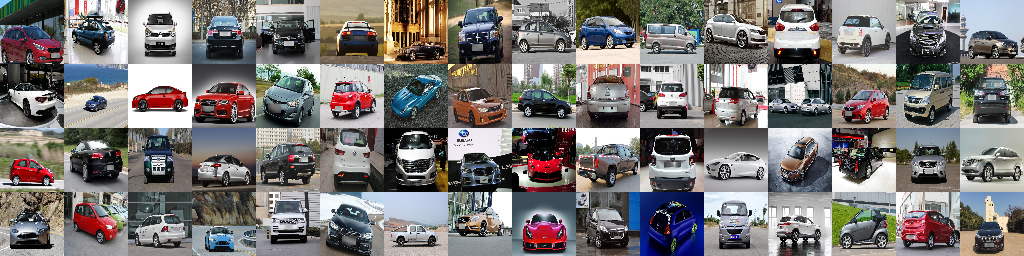}
            \caption{Samples from \compcars{}.}
        \end{figure}

\section{\disc{} Metric Details}
\label{app:metric_details}
    \subsection{Training of Evaluated Models}
        We used \dislib{} \citep{locatello2019challenging}, enabling easy reproducibility of our results, to train all models for \shapes{}, \abdsprites{} and \mpitoy{}.
        We focus on \factorVAE{} and \betaTCVAE{} as they are the models generally achieving the highest disentanglement performance in \citep{locatello2019challenging}.
        Following \citep{locatello2019challenging} on a reduced scale, we trained 5 random seeds each for 6 different hyperparameters.
        This leads to a total of 60 model instances per dataset, spanning a broad range of disentanglement performance.
        
        We followed a similar approach to reproduce the 30 \factorVAE{} and 30 \betaTCVAE{} models for each dataset of the abstract reasoning study \citep{van2019disentangled}. For each VAE model instance, we then trained 5 downstream models on the abstract reasoning task, based on the representation of the VAE, for 6000 steps and evaluated their performance.
        We used \dislib{} to compute the metrics for all evaluated models. 
        
    \subsection{Sampling Strategies for $\tilde{D}_{\text{EP}}$ and $\tilde{D}_{\text{FP}}$}
    \label{app:sampling_strategies}
        To sample from $\tilde{D}_{\text{EP}}$, one can simply sample random instances from the data distribution and then use the model $\mathcal{M}$ to reconstruct these instances.
        This implicitly computes the EP latent distribution $q(z)$.
        To sample from $S_{FP}$, we first need to estimated $\bar{q}(z)$.
        Closely following Algorithm 1 in \citep{kim2018disentangling}, we can achieve this by sampling a batch from $q(z)$ and then randomly permuting across the batch for each latent dimension (see \textit{PermuteDims} in Algorithm \ref{algo:dava}).
        This is a standard trick used in the independence testing literature \citep{arcones1992bootstrap}.
        As long as the batch is large enough, the distribution of these samples samples will closely approximate $\bar{q}(z)$.
        For the evaluation of the \disc{} metric, it is therefore not necessary to compute the marginals over the complete data distribution.
        To amplify the differences between $S_{EP}$ and $S_{FP}$, we then slightly adjusted the sampling strategy for the factorial posterior (FP).
        For each dimension of the latent representation we first estimate its range while encoding a batch of observations.
        We then sample values uniformly in that range, separately for each dimension, resulting in a similar factorial distribution as described earlier, but including a dimension-wise uniform distribution prior.
        This slightly improves performance of the \disc{} metric.
        The architecture of the discriminator used in \disc{} is similar to the encoder of the VAE displayed in Table \ref{tab:models}.

    \subsection{The Role of Reconstruction Loss}
    \label{subsec:app/reconstruction_loss}
\begin{figure}[h!]
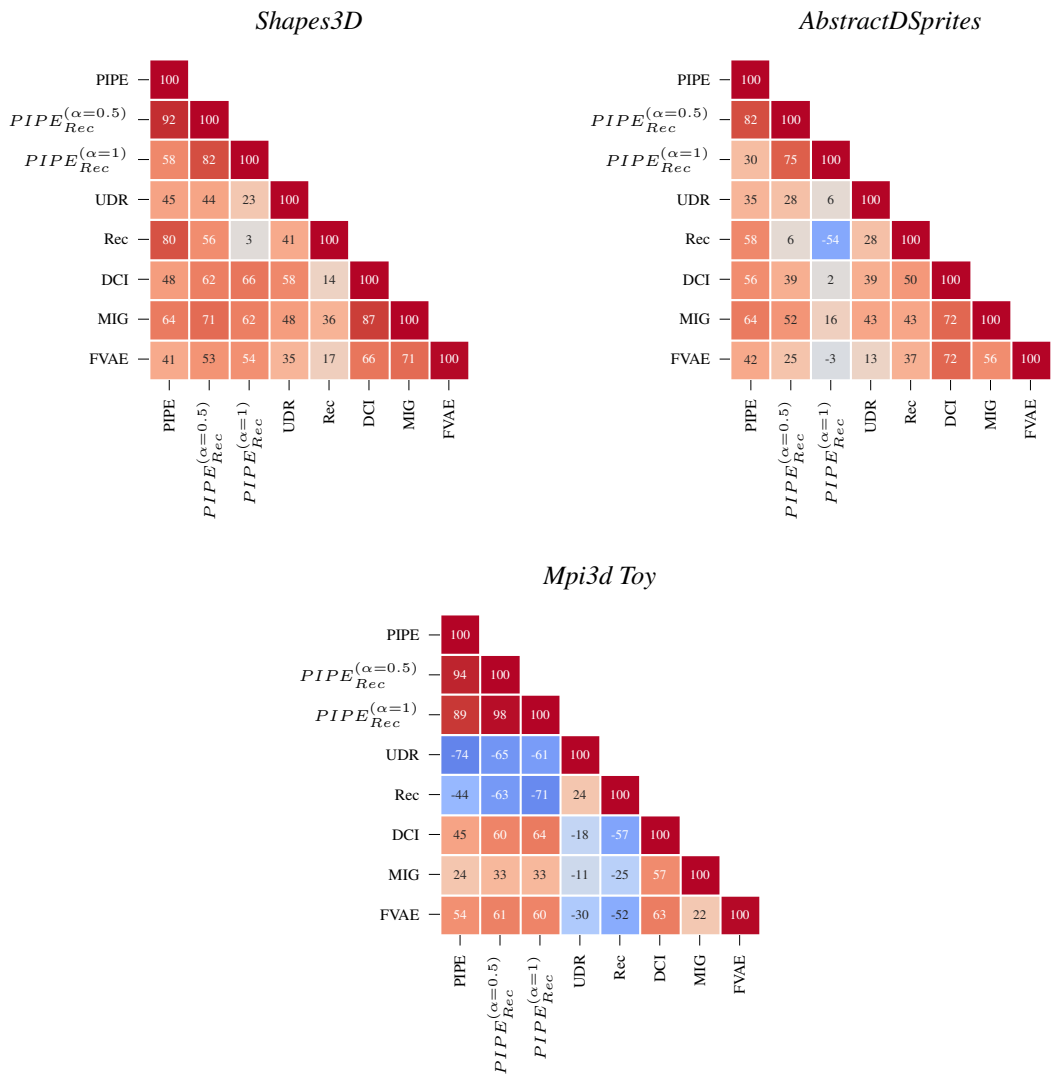

            \centering
            \centering
            \subfloat{\includegraphics[height=7cm]{figures/disc_rec/disc_3DShapes_correlations.tikz}}
            \hfill
            \centering
            \subfloat{\includegraphics[height=7cm]{figures/disc_rec/disc_AbstractDSprites_correlations.tikz}}
            \hfill
            \centering
            \subfloat{\includegraphics[height=7cm]{figures/disc_rec/disc_Mpi3d_toy_correlations.tikz}}
            \caption{Spearman rank correlation between various metrics on three different datasets, similar as Figure \ref{fig:metric/correlation_supervised}. We additionally include $\disc{}_{Rec}$ for $\alpha=0.5$ and $\alpha=1$.}
            \label{fig:app/correlation_supervised_rec}
        \end{figure}

        Consider a model which does not store any information in its latent space, where the decoder outputs the same image regardless of the inputs it receives.
        As the \disc{} metric does not (and cannot) have any knowledge of the ground truth factors, it will give such a model a perfect score\footnote{In fact, this happened with a few \factorVAE{} models on \mpitoy{}, see Table \ref{tab:results_mpitoy}.}. 
        It is possible to use the reconstruction loss as a proxy for how good a model manages to capture the ground truth factors.
        However, the usefulness of the reconstruction loss is limited by two factors.
        First, it is strongly affected by noise, which requires at least dataset-specific normalization for it to be used in a metric.
        Second, the contribution of different ground truth factors in the reconstruction loss can differ a lot. 
        We define \[\disc{}_{Rec} = \disc{} - \alpha \cdot \mathnormal{Rec}_{norm}\] with $\mathnormal{Rec}_{norm}$ denoting the per dataset normalized reconstruction loss.
        We show results for $\alpha=0.5$ and $\alpha=1$ in Figure \ref{fig:app/correlation_supervised_rec}.
        While adding the reconstruction loss brings improved correlations with supervised metrics for  \shapes{} and \mpitoy{}, it decreases correlations for \abdsprites{}, especially for $\alpha=1$.
        For this reason, we decided to propose \disc{} without any reconstruction loss component, as a practitioner, depending on the task, might have different requirements.
        In practice, we suggest to use a classical (denoising) VAE to get a feeling for the achievable reconstruction loss on a dataset and then use PIPE to select the best disentangling models among those that achieve a satisfactory reconstruction loss. 
        
    \subsection{Relation to FID}
\begin{figure}[h!]
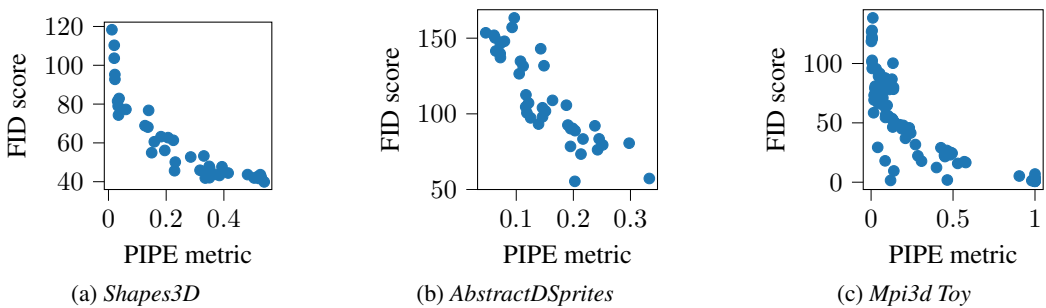

            \centering
            \subfloat[\shapes{}]{\includegraphics[height=3.5cm]{figures/pipe_vs_fid/fid_similar_disc_plot_3dshapes.tikz}}
            \hfill
            \subfloat[\abdsprites{}]{\includegraphics[height=3.5cm]{figures/pipe_vs_fid/fid_similar_disc_plot_abstract_dsprites.tikz}}
            \hfill
            \subfloat[\mpitoy{}]{\includegraphics[height=3.5cm]{figures/pipe_vs_fid/fid_similar_disc_plot_mpi3d_toy.tikz}}
            \caption{FID vs PIPE metric quantifying the similarity between $D_{EP}$ and $D_{FP}$ on three different datasets over all evaluated models. If both metrics would correlate perfectly, one would expect a line from top left to bottom right corner.}
            \label{fig:app/fid_vs_pipe}
        \end{figure}

        The Fréchet inception distance (FID) \citep{heusel2017gans} is a metric to quantify the similarity between two image distributions.
        This is commonly achieved by using a pretrained Inception v3 model \citep{szegedy2016rethinking} without its final classification layer to encode the image distributions. The FID is then the Fréchet distance of two Gaussian distributions fitted to the respective latent image distributions.
        It is commonly used to evaluate the performance of GANs where one desires the generated images to be as similar as possible to the original dataset.
        While it appears that there are strong similarities between FID and \disc{}, there exist distinct differences.
        The contribution of \disc{} is to show the importance of the concept of similarity of the $D_{EP}$ and $D_{FP}$ distributions in light of disentanglement. 
        Compared to our proposed \disc{} metric, FID would be another way to quantify \disc{}.
        Figure \ref{fig:app/fid_vs_pipe} shows that FID and the \disc{} metric have some correlation, while also showing substantial differences for some models.
        A large advantage of the \disc{} metric over FID would be that it is not dependent on a pretrained model.
        This opens up the possibility to deploy the PIPE metric on unseen data domains without any pretrained models available.

    \subsection{Comparison to ModelCentrality}
        Even though MC outperforms UDR, PIPE still considerably outperforms MC. We discuss possible reasons on why this might be the case. 
        To evaluate similarity of two models, MC samples from the prior of a model, but the EP distribution of the model might be far away from the prior. PIPE does not have this problem because it samples from the EP and FP distributions. To evaluate similarity of two models in MC, the second model is asked to encode generated samples of the first model (very similar to $\tilde(D)_{EP}$). 
        This distribution might be far away from the initial data distribution the encoder of the second model is used to.
        Another issue is that if only a very small number of models actually finds a disentangled representation, it is hard for MC to identify these models. PIPE can identify even a single disentangled model among many. Further, due to implicit biases in model architecture, it is possible that models learn similar entangled representations, thus breaking the core assumption of MC that only disentangled representations are similar.

\section{Hyperparameters and \aaae{} Training Details}
    \subsection{Hyperparameters}
        \label{app:hyperparams}
        \aaae{} and all considered baseline models use the same VAE architecture as displayed in Table \ref{tab:models}.
        The architecture of the discriminator and its specific hyperparameters used in \factorVAE{} are shown in Table \ref{tab:fvae_disc}.
        All models, \aaae{} and baseline, use the same general hyperparameters displayed in Table \ref{tab:general_hyperparams}. 
        We present the hyperparameter range explored for each baseline model in Table \ref{tab:param_range}.
        We adjusted the latent space size z\_dim for all models for a small number of datasets.
        We did this in order to give the VAE enough latent capacity to capture the complexity of the dataset.
        More specifically, we set z\_dim to 20 for \mpitoy{} and \mpireal{}, and to 50 for \celeba{} respectively.

        \aaae{} uses the same architecture and general hyperparameters as the baseline models.
        The discriminator of \aaae{} uses the same architecture as the encoder, with the last FC layer having a single output.
        To stabilize the adversarial training procedure of \aaae{}, we had to make use of a few measures more commonly used in GAN training.
        We used label smoothing as well as mixed batches for the training of the discriminator.
        We clipped the gradient norm for both the VAE and the discriminator at 1.
        As specified in Table \ref{tab:app_dava_hyperparams}, we had to use a lower weight for applying the adversarial gradient of the discriminator to the decoder.
        A bigger weight negatively affected reconstruction performance of the VAE.
        We applied instance normalization to the discriminator, which improved the adversarial gradient signal.
        We experimented with other types of normalization (batch, spectral, layer) for encoder, decoder and discriminator, but did not find a consistent improvement for any of them.
        Further, we improved the mechanism to control the KL-Divergence of a VAE proposed by \citep{burgess2018understanding}. Instead of using the absolute difference, we propose to take the difference to the power of 4. This allows the KL-Divergence to fluctuate in between batches, but strongly punishes big deviations.
        
        We summarize the training procedure of \aaae{} in Algorithm \ref{algo:dava}. We use dimension-wise permutation ($\mathnormal{PermuteDims}$) to approximate the factorial posterior FP, similar to \factorVAE{} \citep{kim2018disentangling} and as explained in \ref{app:sampling_strategies}. The process of adjusting $C$ based on the accuracy of the discriminator is described in Algorithm \ref{algo:update_c}.
    
        \begin{table}[H]
              \centering
              \begin{tabular}{ll}
                \toprule
                Encoder / \aaae{} Discriminator     &       Decoder               \\
                \midrule
                Input: [64,64,num channels]        &       FC, 256 ReLU    \\
                4x4 conv, 2 strides, 32 ReLU       &       FC, 4x4x64 ReLU   \\
                4x4 conv, 2 strides, 32 ReLU       &       4x4 upconv, 2 strides, 64 ReLU   \\
                4x4 conv, 2 strides, 64 ReLU       &       4x4 upconv, 2 strides, 64 ReLU   \\
                4x4 conv, 2 strides, 64 ReLU       &       4x4 upconv, 2 strides, 64 ReLU   \\
                FC 256, FC $2 \cdot z\_dim$       &       4x4 upconv, 2 strides, num channels   \\
                \bottomrule
              \end{tabular}
              \caption{VAE architecture used for all models in this study.}
             \label{tab:models}
        \end{table}
        
        \begin{table}[H]
        \centering
            \begin{tabular}{ll}
                \toprule
                Parameter                               &       Value       \\
                \midrule
                Batch size                              &       128 \\
                Optimizer                               &       Adam\\
                Adam: beta1                             &       0.9\\
                Adam: beta2                             &       0.999  \\
                Adam: epsilon                           &       1e-8\\
                Learning rate                           &       0.0001\\
                Training steps                          &       150000 batches\\
                Max. gradient norm                      &       1\\
                \annealedVAE{}: $\gamma$                &       1000\\
                \annealedVAE{}: iteration threshold     &       100000 batches\\
                z\_dim                                  &       10\\
                \bottomrule
              \end{tabular}
              \caption{General hyperparameters.}
              \label{tab:general_hyperparams}
        \end{table}
        
        \begin{table}[H]
              \centering
              \begin{tabular}{ll}
                \toprule
                Parameter                               &       Value       \\
                \midrule
                Optimizer                               &       Adam\\
                Adam: beta1                             &       0.5\\
                Adam: beta2                             &       0.9\\
                Adam: epsilon                           &       1e-8\\
                Learning rate                           &       0.0001\\
                \bottomrule
              \end{tabular}
              \quad
              \quad
              \begin{tabular}{l}
                \toprule
                Discriminator   \\
                \midrule
                FC, 1000 leaky ReLU \\
                FC, 1000 leaky ReLU \\
                FC, 1000 leaky ReLU \\
                FC, 1000 leaky ReLU \\
                FC, 1000 leaky ReLU \\
                FC, 1000 leaky ReLU \\
                FC, 2\\
                \bottomrule
              \end{tabular}
              \caption{Hyperparameters and architecture of the discriminator of \factorVAE{}}
              \label{tab:fvae_disc}
        \end{table}
        
        \begin{table}[H]
              \centering
              \begin{tabular}{lll}
                \toprule
                Model   &           Parameter   &           Value       \\
                \midrule
                \betaTCVAE{} &      $\beta$     &           $[1, 2, 4, 8, 16, 32]$\\
                \factorVAE{} &      $\gamma$    &           $[5, 10, 20, 30, 50, 100]$\\
                \annealedVAE{} &    $C$         &           $[\frac{1}{2}, 1, 2, 5, 25, 50]$\\
                \CVAE{} &           $C$         &           $[16, 18, 25, 35, 180, 200]$\\
                \bottomrule
              \end{tabular}
              \caption{Hyperparameter range explored for each model}
              \label{tab:param_range}
        \end{table}

        \begin{table}[H]
              \centering
            \begin{tabular}{ll}
                \toprule
                Parameter                               &       Value       \\
                \midrule
                Discriminator: Optimizer                               &       Adam\\
                Discriminator: Adam: beta1                             &       0.9\\
                Discriminator: Adam: beta2                             &       0.999\\
                Discriminator: Adam: epsilon                           &       1e-8\\
                Discriminator: Learning rate                           &       0.0001\\
                Discriminator: Max. gradient norm                      &       1\\
                \midrule
                Dis encoder weight $\mu_{enc}$      &       0.3\\
                Dis decoder weight $\mu_{dec}$     &       0.001\\
                $\gamma$                &       500\\
                $\Delta C$              &       4e-5\\
                \bottomrule
              \end{tabular}
              \caption{\aaae{} hyperparameters}
                \label{tab:app_dava_hyperparams}
        \end{table}

         \begin{algorithm}
            \begin{algorithmic}
                \State $\theta_{enc}, \theta_{dec}, \theta_{dis} \gets \text{initialize parameters}$
                \While{Not Converged}
                    \State $x \gets \text{sample random mini-batch from } \tilde{D}$ 
                    \State $\hat{z} \gets Enc(x)$
                    \State $\hat{x} \gets Dec(\hat{z})$
                    \State $\mathcal{L}_{vae} \gets \mathrm{log} p( \hat{x}| \hat{z})
                            - \gamma (\mathrm{KL}(q(\hat{z})||p(\hat{z})) - C)^4 $
                    \State $\theta_{enc} \stackrel{+}\gets -\nabla_{\theta_{enc}}\mathcal{L}_{vae}$
                    \State $\theta_{dec} \stackrel{+}\gets -\nabla_{\theta_{dec}}\mathcal{L}_{vae}$
                    \Comment Update VAE
                    \State $\hat{z} \gets Enc(x)$
                    \State $\hat{x} \gets Dec(\hat{z})$
                    \Comment Recreate $\hat{z}$ and $\hat{x}$ after update
                    \State $\tilde{x} \gets Dec(\mathnormal{PermuteDims}(\hat{z}))$
                    \State $acc \gets \mathnormal{accuracy}(Dis, \hat{x}, \tilde{x})$
                    \State $\theta_{dis} \stackrel{+}\gets -\nabla_{\theta_{dis}}\mathrm{log}(Dis(\hat{x})) +  \mathrm{log}(1 - Dis(\tilde{x}))$
                    \Comment Update Discriminator
                    \State $C \stackrel{+}\gets \mathnormal{update_C}(acc)$
                    \State $\mu_{base} \gets max((acc - 0.5) \cdot 100, 0)$
                    \Comment Weight grows linearly with higher accuracy
                    \State $\theta_{enc} \stackrel{+}\gets -\nabla_{\theta_{enc}} \mu_{base} \cdot \mu_{enc} \cdot \mathrm{log}(Dis(\hat{x}))$
                    \State $\theta_{dec} \stackrel{+}\gets -\nabla_{\theta_{dec}} \mu_{base} \cdot \mu_{dec} \cdot \mathrm{log}(1 - Dis(\hat{x}))$
                    \Comment Update VAE with adversarial loss
                \EndWhile
             \end{algorithmic}
             \caption{Training \aaae{}}
             \label{algo:dava}
        \end{algorithm}
        
        \begin{algorithm}
           \begin{algorithmic}
               \State \textbf{Input:} accuracy $acc$
               \If{$acc \leq 0.5$}
                    \State $U \gets \Delta C$
                    \Comment Only increase $C$ if discriminator is clueless
                \ElsIf{$acc  \leq 0.51$}
                    \State $U \gets 0$
                    \Comment Grace period
                \Else
                    \State $U \gets - \Delta C$
                    \Comment Decrease $C$ if discriminator gets better
                \EndIf
               \State \textbf{Output:} update $U$
           \end{algorithmic}
           \caption{$\mathnormal{update_C}$}
           \label{algo:update_c}
        \end{algorithm}

    \subsection{Individual Contribution of Different Building Blocks of \aaae{}}
        One might wonder which building blocks of \aaae{} contribute most to the performance.
        Results during development show that the adversarial loss of \aaae{} alone delivers comparable performance to \betaTCVAE{} and \factorVAE{} with the same issue of regularization strength being dependent on the dataset.
        Interestingly, for \annealedVAE{}, the optimal $C$ for different datasets closely correspond to the values discovered by \aaae{} displayed in Figure \ref{fig:training/capacity}.
        This also reflects the performance one could expect from \aaae{} without the adversarial loss.
        
    \subsection{Combination of \aaae{} with Existing Methods}
        The main contribution of \aaae{} lies in its ability to adapt the latent capacity bottleneck with respect to the current state of the model, evaluated by the discriminator.
        In that regard, it would be possible to use a different discriminator as the one proposed by \aaae{}.
        One such option is the discriminator used in \factorVAE{}, which could be used in a similar manner as the discriminator in \aaae{} to guide the latent capacity bottleneck.
        This would still allow for a dataset-specific annealing scheme without the need for dataset-specific hyperparameters.

\section{Complete Quantitative Results of all Evaluated Datasets}
\label{app:complete_results}
    We present results for a large range of datasets.
    We also include results for the best regularization strength for each of the baseline approaches after supervised hyperparameter selection in the upper part of each table.
    This serves as an additional reference on what can be expected from the completely unsupervised approaches in a best case scenario.
    We further include the results of Recursive Disentanglement Network (RecurD) by \citet{chen2021recursive}.
    These are to be interpreted with a grain of salt, as we were not able to reproduce them ourselves, but had to take the values from their paper.
    This means that first, only a limited set of datasets and metrics were evaluated.
    Second, there will be differences in model architecture, hyperparameter choice, the range of hyperparameter tuning, general training loop and possibly more things.
    Please note that there are two aspects to take into consideration in regard of the strong performance of \annealedVAE{}.
    First, the selected range of hyperparameters is especially strong, as we include the hyperparameters discovered by \aaae{}.
    Second, as explained earlier for \aaae{}, we use $(\cdot)^4$ instead of the absolute value of the difference for controlling the KL-divergence.
    Further note that the exceptionally high \disc{} metric score of \factorVAE{} on \textit{NoisyDSprites} is caused by the decoder of the \factorVAE{} models always producing the same output regardless of the latent space.
    This is a design decision of \disc{}, as described in Subsection \ref{subsec:app/reconstruction_loss}.
    Perhaps surprisingly, these models still seem to convey a small amount of information in the latent space as reflected by the supervised metrics.
    The same occurred with \factorVAE{} for \mpitoy{}, interestingly though not on \mpireal{}.
    
     \begin{table}[H]
      \caption{Results \shapes{}}
      \label{tab:complete_results_shapes}
        \centering
        \begin{tabular}{ l c c c c c }
            \toprule
            Architecture &                          MIG &                   DCI &                   FVAE &              PIPE &              Rec \\
            \midrule
            Best \betaTCVAE{} ($\beta=32$)&         0.39±0.09 &             0.65±0.03  &            0.76±0.06 &         0.36±0.03 &         0.0041±0.0008 \\
            Best \factorVAE{} ($\gamma=30$)&        0.39±0.16 &             0.54±0.12  &            0.79±0.05 &         0.30±0.08 &         0.0032±0.0005\\
            Best \annealedVAE{} ($C=1$) &           \bw{0.63±0.04} &        \bw{0.78±0.05} &        \bw{0.94±0.04} &    \bw{0.49±0.05} &    0.0027±0.0002\\
            Best \CVAE{} ($C=18$) &                 0.08±0.03 &             0.20±0.06 &             0.68±0.12 &         0.02±0.00 &         \bw{0.0019±0.0004} \\
            Best RecurD \citep{chen2021recursive}&   (0.31) &                (0.58)  &               - &                 - &                 (0.0083)\\
            \midrule
            Mean \betaTCVAE{} &                     0.26±0.06 &             0.50±0.03 &             0.76±0.03 &         0.11±0.03 &         0.0023±0.0004\\
            Mean \factorVAE{} &                     0.20±0.10 &             0.41±0.07 &             0.78±0.04 &         0.22±0.04 &         0.0027±0.0005\\
            Mean \annealedVAE{} &                   0.52±0.02 &             0.70±0.02 &             \bo{0.92±0.01} &    0.26±0.03 &         0.0020±0.0001\\
            Mean \CVAE{} &                          0.07±0.03 &             0.17±0.04 &             0.68±0.07 &         0.02±0.00 &         \bo{0.0019±0.0004} \\
            Ours &                                  \bo{0.62±0.05} &        \textbf{0.78±0.03} &    0.82±0.03 &         \textbf{0.61±0.04}& 0.0029±0.0004 \\
            \bottomrule
        \end{tabular}
    \end{table}
    
    \begin{table}[H]
      \caption{Results \abdsprites{}}
      \label{tab:complete_results_abdsprites}
        \centering
        \begin{tabular}{ l c c c c c }
            \toprule
            Architecture &                      MIG &                   DCI &                   FVAE &                  PIPE &              Rec \\
            \midrule
            Best \betaTCVAE{} ($\beta=4$) &     0.17±0.04 &             0.22±0.03 &             0.55±0.03 &             0.11±0.02 &         0.0017±0.0001\\
            Best \factorVAE{} ($\gamma=50$)&    0.18±0.07 &             \bw{0.26±0.04} &        \bw{0.65±0.05} &        \bw{0.23±0.04} &    0.0019±0.0001\\
            Best \annealedVAE{} ($C=2$) &       \bw{0.23±0.04} &        \bw{0.26±0.04} &        \bw{0.65±0.05} &        0.13±0.02   &       0.0018±0.0001\\
            Best \CVAE{} ($C=18$) &             0.06±0.01 &             0.11±0.03 &             0.50±0.05 &             0.07±0.01 &         \bw{0.0016±0.0000} \\
            \midrule
            Mean \betaTCVAE{} &                 0.12±0.01 &             0.18±0.01&              0.45±0.02 &             0.19±0.01 &         0.0027±0.0000\\
            Mean \factorVAE{} &                 0.12±0.02 &             0.19±0.02 &             0.55±0.03 &             0.20±0.03 &         0.0018±0.0000\\
            Mean \annealedVAE{} &               0.15±0.01 &             0.21±0.01 &             0.54±0.02 &             0.18±0.01 &         0.0022±0.0000\\
            Mean \CVAE{} &                      0.04±0.02 &             0.07±0.01 &             0.42±0.02 &             0.06±0.01 &         \bo{0.0016±0.0000} \\
            Ours &                              \textbf{0.23±0.04} &    \textbf{0.27±0.05} &    \textbf{0.67±0.05} &    \textbf{0.35±0.03}& 0.0020±0.0002 \\
            \bottomrule
        \end{tabular}
    \end{table}
    
    \begin{table}[H]
      \caption{Results  \mpitoy{}}
      \label{tab:complete_results_mpitoy}
        \centering
        \begin{tabular}{ l c c c c c }
            \toprule
             Architecture &                         MIG &                   DCI &                   FVAE &                  PIPE &          Rec \\
            \midrule
            Best \betaTCVAE{} ($\beta=2$)&          \bw{0.23±0.04} &        \bw{0.33±0.01} &        0.42±0.06 &             0.07±0.01 &     0.0005±0.0000\\
            Best \factorVAE{} ($\gamma=5$)&         0.10±0.05   &           0.20±0.02 &             \bw{0.46±0.03} &        \bw{0.36±0.16}& 0.0021±0.0003 \\
            Best \annealedVAE{} ($C=\frac{1}{2}$) & 0.21±0.07 &             0.31±0.02 &             0.41±0.05 &             0.12±0.01 &     0.0005±0.0000\\
            Best \CVAE{}($C=16$) &                  0.04±0.01 &             0.18±0.02 &             0.44±0.04 &             0.02±0.01 &     \bw{0.0003±0.0000} \\
            \midrule
            Mean \betaTCVAE{} &                     0.11±0.02 &             0.23±0.01 &             0.39±0.02 &             0.09±0.01 &     0.0006±0.0000 \\
            Mean \factorVAE{} &                     0.02±0.01 &             0.13±0.00 &             0.38±0.02 &             0.99±0.10  &    0.0041±0.0000   \\
            Mean \annealedVAE{} &                   0.07±0.03 &             0.23±0.02 &             \textbf{0.50±0.02} &    0.02±0.01 &     \bo{0.0003±0.0000}   \\
            Mean \CVAE{} &                          0.04±0.01 &             0.17±0.02 &             0.43±0.04 &             0.03±0.01 &     \bo{0.0003±0.0000} \\
            Ours &                                  \bo{0.12±0.09} &        \bo{0.30±0.03} &        0.41±0.04 &             \bo{0.21±0.03}& 0.0006±0.0000  \\
            \bottomrule
        \end{tabular}
    \end{table}

    \begin{table}[H]
      \caption{Results \dsprites{}}
      \label{tab:complete_results_dsprites}
        \centering
        \begin{tabular}{ l c c c c c }
            \toprule
            Architecture &                          MIG &                   DCI &                   FVAE &                  PIPE &          Rec\\
            \midrule
            Best \betaTCVAE{} ($\beta=32$) &		0.34±0.02 &	            0.50±0.03 &	            0.60±0.00 &	            \bw{0.76±0.14}& 0.0081±0.0000 \\
            Best \factorVAE{} ($\gamma=50$) &		0.10±0.04 &	            0.22±0.02 &	            0.73±0.04 &	            0.41±0.05 &     0.0098±0.0007\\
            Best \annealedVAE{} ($C=\frac{1}{2}$) &	\bw{0.36±0.01} &	    \bw{0.51±0.01} &	    \bw{0.76±0.03} &	    0.42±0.03 &     0.0060±0.0006\\
            Best \CVAE{} ($C=18$) &                 0.06±0.02 &             0.13±0.03 &             0.67±0.11 &             0.24±0.03 &     \bw{0.0008±0.0000} \\
            Best RecurD \citep{chen2021recursive} &  (0.27)    &             (0.38) &                - &                     - &             (0.0047)\\
            \midrule
            Mean \betaTCVAE{} & 			        \bo{0.27±0.05} &	    \bo{0.45±0.06} &	    \bo{0.84±0.03} &	    0.44±0.08&      0.0024±0.0001\\
            Mean \factorVAE{} & 			        0.14±0.06 &	            0.22±0.05 &	            0.72±0.07 &	            0.31±0.03 &     0.0052±0.0002\\
            Mean \annealedVAE{} & 			        0.18±0.06 &	            0.28±0.05 &	            0.77±0.04 &	            0.32±0.05 &     0.0013±0.0001\\
            Mean \CVAE{} &                          0.05±0.02 &             0.13±0.01 &             0.69±0.06 &             0.20±0.08 &     \bo{0.0008±0.0000}\\
            Ours & 	 		                        0.24±0.09 &	            0.42±0.14 &	            0.74±0.11 &	            \bo{0.51±0.09}& 0.0025±0.0005\\
            \bottomrule
        \end{tabular}
    \end{table}
    
    \begin{table}[H]
      \caption{Results \mpireal{}}
      \label{tab:complete_results_mpireal}
        \centering
        \begin{tabular}{ l c c c c c }
            \toprule
            Architecture &                      MIG &                   DCI &                   FVAE &                  PIPE &          Rec\\
            \midrule
            Best \betaTCVAE{} ($\beta=2$) &		\bw{0.16±0.08} &	    \bw{0.35±0.06} &	    \bw{0.56±0.05} &	    0.03±0.01&      0.0004±0.0000 \\
            Best \factorVAE{} ($\gamma=30$) &	0.02±0.01 &	            0.18±0.04 &	            0.43±0.03 &	            \bw{0.42±0.10}& 0.0031±0.0003 \\
            Best \annealedVAE{} ($C=1$) &		0.12±0.08 &	            0.30±0.07 &	            0.49±0.08 &	            0.08±0.01 &     0.0004±0.0000\\
            Best \CVAE{} ($C=25$) &             0.06±0.02 &             0.22±0.01 &             0.54±0.05 &             0.01±0.00 &     \bw{0.0003±0.0000} \\
            \midrule
            Mean \betaTCVAE{} & 			    0.08±0.04 &	            0.22±0.03 &	            0.26±0.03 &	            \bo{0.23±0.07}& 0.0015±0.0001 \\
            Mean \factorVAE{} & 			    0.05±0.01 &	            0.19±0.01 &	            0.42±0.02 &	            0.12±0.07 &     0.0022±0.0003\\
            Mean \annealedVAE{} & 			    0.09±0.03 &	            \bo{0.29±0.02} &	    \bo{0.52±0.05} &	    0.03±0.01 &     \bo{0.0003±0.0000}\\
            Mean \CVAE{} &                      0.04±0.02 &             0.20±0.03 &             0.52±0.06 &             0.01±0.00 &     \bo{0.0003±0.0000} \\
            Ours & 	 		                    \bo{0.11±0.05} &	    0.27±0.03 &	            0.48±0.05 &	            0.20±0.04 &     0.0006±0.0000\\
            \bottomrule
        \end{tabular}
    \end{table}
    
    \begin{table}[H]
      \caption{Results \textit{Smallnorb}}
      \label{tab:complete_results_smallnorb}
        \centering
        \begin{tabular}{ l c c c c c }
            \toprule
            Architecture &                          MIG &                   DCI &                   FVAE &                  PIPE &          Rec \\
            \midrule
            Best \betaTCVAE{} ($\beta=4$) &		    0.16±0.03 &	            \bw{0.34±0.01} &	    0.60±0.02 &	            0.18±0.03&      0.0037±0.0000 \\
            Best \factorVAE{} ($\gamma=5$) &		0.12±0.06 &	            0.25±0.03 &	            \bw{0.64±0.02} &	    \bw{0.24±0.03}& 0.0054±0.0009 \\
            Best \annealedVAE{} ($C=\frac{1}{2}$) &	\bw{0.20±0.01} &	    0.28±0.01 &	            0.58±0.01 &	            0.22±0.02&      0.0037±0.0000 \\
            Best \CVAE{} ($C=180$) &                0.24±0.01 &             0.28±0.01 &             0.62±0.04 &             0.03±0.00 &     \bw{0.0016±0.0001} \\
            \midrule
            Mean \betaTCVAE{} & 			        0.18±0.02 &	            0.29±0.01 &	            0.57±0.02 &	            0.11±0.01 &     0.0032±0.0001\\
            Mean \factorVAE{} & 			        0.03±0.03 &	            0.13±0.01 &	            0.53±0.04 &	            0.17±0.09 &     0.0101±0.0039\\
            Mean \annealedVAE{} & 			        0.21±0.01 &	            0.28±0.00 &	            0.60±0.01 &	            0.05±0.00 &     0.0020±0.0001\\
            Mean \CVAE{} &                          0.23±0.02 &             0.28±0.01 &             0.62±0.03 &             0.03±0.00 &     \bo{0.0016±0.0001} \\
            Ours & 	 		                        \bo{0.25±0.01} &	    \bo{0.30±0.01} &	    \bo{0.62±0.01} &       \bo{0.25±0.02}&  0.0030±0.0001 \\
            \bottomrule
        \end{tabular}
    \end{table}
    
    \begin{table}[H]
      \caption{Results \textit{NoisyDSprites}}
      \label{tab:complete_results_noisydsprites}
        \centering
        \begin{tabular}{ l c c c c c }
            \toprule
            Architecture &                      MIG &                   DCI &                   FVAE &                  PIPE &          Rec \\
            \midrule
            Best \betaTCVAE{} ($\beta=4$) &		\bw{0.08±0.03} &	    \bw{0.21±0.04} &	    0.63±0.09 &	            0.32±0.01 &     0.0804±0.0000\\
            Best \factorVAE{} ($\gamma=100$) &	0.01±0.00 &	            0.10±0.05 &	            0.38±0.09 &	            \bw{1.00±0.00}& 0.0892±0.0000 \\
            Best \annealedVAE{} ($C=50$) &		0.06±0.02 &	            0.13±0.03 &	            \bw{0.65±0.08} &	    0.48±0.29 &     0.0802±0.0000\\
            Best \CVAE{} ($C=16$) &             0.02±0.01 &             0.08±0.01 &             0.48±0.06 &             0.43±0.29 &     \bw{0.0801±0.0000} \\
            \midrule
            Mean \betaTCVAE{} & 			    0.05±0.03 &	            0.11±0.05 &	            0.44±0.08 &	            0.31±0.05 &     0.0810±0.0000\\
            Mean \factorVAE{} & 			    0.01±0.00 &	            0.06±0.02 &	            0.33±0.03 &	            (1.00±0.00)&    0.0892±0.0000 \\
            Mean \annealedVAE{} & 			    \bo{0.09±0.05} &	    0.18±0.06 &	            \bo{0.58±0.09} &	    0.27±0.03  &    0.0805±0.0000 \\
            Mean \CVAE{} &                      0.02±0.01 &             0.08±0.01 &             0.49±0.01 &             0.35±0.14 &     \bo{0.0801±0.0000} \\
            Ours & 	 		                    \bo{0.09±0.05} &	    \bo{0.22±0.05} &	    0.55±0.07 &             \bo{0.32±0.04}& 0.0806±0.0001 \\
            \bottomrule
        \end{tabular}
    \end{table}
    
    \begin{table}[H]
      \caption{Results \textit{Cars3D}}
      \label{tab:complete_results_cars3d}
        \centering
        \begin{tabular}{ l c c c c c }
            \toprule
            Architecture &                          MIG &                   DCI &                   FVAE &                  PIPE &          Rec\\
            \midrule
            Best \betaTCVAE{} ($\beta=16$) &        \bw{0.14±0.02} &	   \bw{0.33±0.08} &	        \bw{0.90±0.02} &	    0.07±0.01 &     0.0059±0.0000 \\
            Best \factorVAE{} ($\gamma=20$) &	    0.10±0.01 &	            0.16±0.04 &	            0.88±0.03 &	            0.10±0.02 &     0.0083±0.0005\\
            Best \annealedVAE{} ($C=1$) &		    0.12±0.01 &	            0.29±0.05 &	            0.89±0.02 &	            \bw{0.15±0.01}& 0.0059±0.0001 \\
            Best \CVAE{} ($C=16$) &                 0.05±0.02 &             0.10±0.02 &             0.84±0.05 &             0.03±0.00 &     \bw{0.0045±0.0000} \\
            Best RecurD \citep{chen2021recursive}&   (0.17) &                - &                     - &                     - &             (0.0132) \\
            \midrule
            Mean \betaTCVAE{} & 			        0.11±0.03 &	            0.24±0.01 &	            0.91±0.02 &	            0.06±0.01 &     0.0049±0.0001 \\
            Mean \factorVAE{} & 			        0.06±0.02 &	            0.12±0.03 &	            0.74±0.17 &	            \bo{0.28±0.14}& 0.0170±0.0023 \\
            Mean \annealedVAE{} & 			        0.06±0.02 &	            \bo{0.25±0.04} &	    0.86±0.01 &	            0.08±0.01 &     0.0047±0.0001\\
            Mean \CVAE{} &                          0.03±0.01 &             0.11±0.01 &             0.76±0.09 &             0.03±0.00 &     \bo{0.0045±0.0000}\\
            Ours & 	 		                        \bo{0.15±0.01} &	    0.23±0.04 &	 	        \bo{0.94±0.01} &        \bo{0.28±0.02}& 0.0074±0.0001\\
            \bottomrule
        \end{tabular}
    \end{table}

\section{Further Results on Real-life Datasets}
\label{app:compcars_results}
\begin{figure}[h!]
            \centering
            \includegraphics[width= 0.5\textwidth]{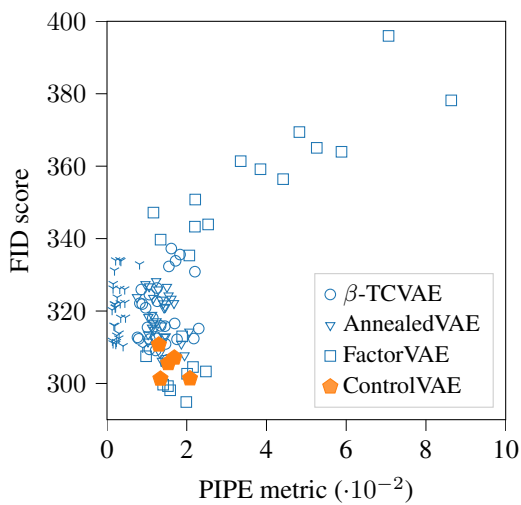}
            \caption{Similar to Figure \ref{fig:real_world/celeba_fid_vs_disc}, we display the FID of image samples from $\tilde{D}_{\text{FP}}$ of the respective models versus the original images of \compcars{}. \aaae{} achieves competitive FID scores even compared to models exploring a broad range of hyperparameters. \factorVAE{} models that achieve better disentanglement according to the PIPE metric suffer from significantly lower FID scores.}
            \label{fig:compcars_fid_plot}
\end{figure}

We further report results of \aaae{} against the full range of baseline models on the \compcars{} dataset. 
As described in the dataset description in \ref{app:compcars}, \compcars{} is an extremely challenging dataset. 
Performance is most likely limited by the model architecture used, as can also be seen by the bad FID and \disc{} metric scores displayed in Figure \ref{fig:compcars_fid_plot}.

\end{document}